\documentclass[preprint,11pt]{elsarticle}
\usepackage{booktabs}

\usepackage{amssymb}
\usepackage{amsmath}
\usepackage{hyperref} 
\usepackage{etex}
\usepackage[utf8]{inputenc}
\usepackage[T5]{fontenc}

\usepackage{xcolor}

\journal{arXiv}

\begin{document}

\begin{frontmatter}



\title{SpiS-GAN: Spiral-Modulated Handwriting Synthesis with Star Operation} 

%

\author[1,2]{Nguyen Duy Hieu} 
\author[1,2]{Dang Hoai Nam}
\author[1,2]{Pham Hoang Giap}
\author[3]{Quang Huu Hieu}
\author[1,2,4]{Vo Nguyen Le Duy \corref{cor1}}

\affiliation[1]{organization={University of Information Technology}, 
	city={Ho Chi Minh City}, 
	country={Vietnam}}

\affiliation[2]{organization={Vietnam National University}, 
	city={Ho Chi Minh City}, 
	country={Vietnam}}

\affiliation[3]{organization={AJ Technologies}, 
	city={Nagoya}, 
	country={Japan}}

\affiliation[4]{organization={RIKEN Center for Advanced Intelligence Project}, 
	city={Tokyo}, 
	country={Japan}}

\cortext[cor1]{Corresponding author. E-mail: duyvnl@uit.edu.vn}      

\begin{abstract}
Training robust handwriting recognition (HTR) systems requires massive amounts of annotated data, which is often difficult to acquire. While synthetic handwriting generation offers a practical solution to expand training sets, existing models struggle with several core issues. First, previous approaches, even MLP-based models fail to effectively trace cursive handwriting due to fixed-grid spatial receptive field. Second, their CNN-relied discriminators usually lose structural details through aggressive downsampling, making broken connections difficult to detect. Third, existing architectures are either limited to linear feature interactions or too expensive for high-resolution synthesis. Finally, existing approaches lack explicit edge constraints, often resulting in blurred stroke boundaries. To address these challenges, this study proposes a Spiral-Modulated Handwriting Synthesis framework based on Generative Adversarial Networks (SpiS-GAN). Our generator employs Star-Spiral Blocks combining proposed Modulated Elliptical SpiralFC with the star operation to capture spatial relationships and efficiently follow complex handwriting stroke trajectories, while a Spiral-Modulated discriminator is introduced for multi-domain flaws detection. Additionally, we introduce a Sobel-Regularized Edge Reconstruction Loss that provides edge guidance, ensuring every character remains clear and legible. Evaluations on the English and Vietnamese datasets demonstrate that SpiS-GAN significantly outperforms current state-of-the-art models. The generated images are highly authentic, accurately preserve the original writer's style across languages, and successfully lower error rates when training downstream HTR systems.

\end{abstract}




\end{frontmatter}



\section{Introduction}\label{sec:introduction}
Even though screens are everywhere, pen and paper are not going away anytime soon. From students taking exams to historians examining centuries-old documents, handwritten text remains a vital part of our daily lives. Beyond just being useful, a person’s handwriting is as unique as their fingerprint, carrying a personal touch that typed fonts simply cannot copy. Because we still rely so much on paper, we need software that can smoothly translate handwritten notes into digital text, a technology called Handwriting Recognition (HTR). However, getting a computer to actually do this in the real world is a massive headache. A reliable HTR system has to decode messy handwriting, faded ink, blurry scans, and confusing grammar rules all at the same time \cite{HANDS-VNOnDB, kleber2013cvl, ICFHR2018, NabucoLatin}. This problem is especially painful for low-resource languages. For these languages, nobody has spent the millions of hours or dollars needed to build the giant digital libraries required to teach an AI how to read them.

Modern AI is incredibly smart, but it is also extremely reliant on massive datasets. If you do not provide it with enough examples, it simply fails to learn. To get around this bottleneck, researchers have started using Handwriting Synthesis (HS), which basically means teaching a computer to generate fake, realistic handwriting to artificially grow the AI's training data \cite{graves2013generating, bhunia2021handwriting}. A perfect HS system takes typed words and draws them so naturally that they look exactly like a specific human wrote them. But teaching a machine to simulate human nuances is incredibly difficult. The AI has to figure out how a specific person connects their letters, spaces their words, and changes their pen strokes, often by looking at just a few small sample images. This task goes from difficult to extreme when generating text in languages like Vietnamese. In Vietnamese, tiny accent marks and stacked tones completely change a word's meaning. Getting a computer to correctly place these layered, complex marks while maintaining the natural, messy flow of human handwriting remains a huge challenge.

While recent Generative Adversarial Networks (GANs) have improved synthesis realism, some core limitations still hinder their performance. First, previous modules attempt to expand receptive fields via criss-cross token mixing along rigid axes, but this ineffectively capture the fluid, looping movements inherent in handwriting. Second, most discriminators rely on CNNs that discard structural details through downsampling, while Transformer-based alternatives are costly. Third, conventional architectures cannot effectively capture long-range stroke relationships without excessively expanding the network. Finally, existing frequency-aware supervision is insufficient for high-frequency information such as sharp stroke boundaries, often resulting in over-smoothed outputs.

To overcome aforementioned limitations, we introduce SpiS-GAN, a spiral-modulated, one-shot handwriting synthesis framework designed to produce realistic and writer-consistent handwritten text from a single reference. Our generator is built upon Star-Spiral Block (SSB) that combine Modulated Elliptical SpiralFC (MESpiralFC) with a pointwise Star product, enabling efficient spatial fusion and stroke trajectory modeling. Furthermore, we introduce a novel Sobel-Regularized Reconstruction Loss (SELoss) that enforces directional edge constraints, ensuring faithful stroke reconstruction. Additionally, we propose an Spiral-Modulated Discriminator that efficiently detects errors without severe spatial reduction.

Experiments on both English and Vietnamese handwriting datasets have verified the excellent performance of the proposed model. Its ability to augment low-resource training data positions it as a practical solution for enhancing HTR systems used in real-world expert applications. The main contributions of this paper are as follows:
\begin{itemize}
	\item We propose a novel Modulated Elliptical SpiralFC's sampling with the star operation's implicit high-dimensional expansion, enabling rich feature interactions in generator.
	\item We introduce a novel edge-aware loss that ensures generated strokes remain clean and well-defined.
	\item We integrate a modified SpiralMLP block into the discriminator, allowing to evaluate features through multiple parallel pathways.
	\item Experiments on English and Vietnamese handwritten datasets demonstrate that SpiS-GAN significantly outperforms the state-of-the-art. Our code is available at:  
	\begin{center}
		\href{https://github.com/DAIR-Group/SpiS-GAN}{https://github.com/DAIR-Group/SpiS-GAN}
	\end{center}
\end{itemize}

The remainder of this paper is organized as follows. Section~\ref{sec:related} reviews related studies on handwriting generation and style transfer. Section~\ref{sec:proposal} details the proposed SpiS-GAN architecture and its objective functions. Section~\ref{sec:experiments} describes the experimental setup and analyzes both quantitative and qualitative results. Finally, Section~\ref{sec:conclusion} concludes the paper and discusses future directions.

\section{Related Work}\label{sec:related}

The field of handwriting text generation (HTG) includes a broad range of techniques designed to produce synthetic handwritten content that closely resembles human writing in both appearance and style. These methods generally fall into two main categories based on how they represent handwriting: online approaches, which simulate the dynamic movement of the pen over time, and offline approaches, which treat handwriting as a static visual image. Each paradigm has its own strengths and limitations.

\subsection{Online Handwriting Generation}

Online HTG methods create handwriting by modeling the pen's trajectory over time. These techniques typically use sequence-based models such as LSTMs~\cite{graves2013generating}, conditional variational recurrent networks~\cite{aksan2018deepwriting}, or temporal CNNs~\cite{aksan2018stcn} to predict a sequence of pen coordinates based on the target text. The pioneering work by Graves~\cite{graves2013generating} established this framework but did not include any mechanism for style conditioning. Later studies~\cite{aksan2018deepwriting, aksan2018stcn, kotani2020generating} addressed this limitation by extracting style information from reference images and incorporating it into the sequence generation process.

Generative Adversarial Networks (GANs) have also been applied to online handwriting generation. For instance, Ji et al.~\cite{ji2019generative} introduced a discriminator to improve the realism of pen stroke trajectories. Despite these advances, online methods still face significant obstacles. First, capturing long-range dependencies in pen motion remains computationally demanding. Second, and perhaps more critically, online data requires temporal stroke recordings, which are costly to collect and simply do not exist for historical manuscripts or standard scanned document collections. As a result, many recent works, including ours, have shifted their focus to the offline generation setting, where data is more readily available and real-world applications are more practical.

\subsection{Offline Handwriting Generation}
\label{sec:offline_handwriting_generation}
Generating handwriting offline means creating the final text as static pictures. In the early days, researchers had to manually isolate letter shapes and write strict programming rules to connect strokes, arrange layouts, and blend backgrounds \cite{wang2005combining, lin2007style, thomas2009synthetic, haines2016my}. This manual work took a massive amount of time and completely broke down when the system encountered new letters or unfamiliar writing styles. Today, artificial intelligence has largely replaced those older manual systems, particularly through the use of Generative Adversarial Networks (GANs) \cite{alonso2019adversarial, fogel2020scrabblegan, kang2020ganwriting, gan2021higan}. For instance, Alonso and colleagues \cite{alonso2019adversarial} built a system that turned text data into word images of a strict, fixed size. The ScrabbleGAN project \cite{fogel2020scrabblegan} took this a step further by stitching small image patches together, allowing the AI to generate text of any length. Later studies \cite{kang2020ganwriting, gan2021higan, mattick2021smartpatch} took similar paths, usually guiding the generation process using specific text encoding methods. Some alternative approaches \cite{luo2022slogan, krishnan2021textstylebrush} bypass this entirely by directly applying a new handwriting style over standard typed images. Furthermore, the way modern systems learn a writing style can differ greatly. Some analyze entire paragraphs \cite{davis2020text}, while others look closely at individual words or even a single reference example \cite{gan2021higan, luo2022slogan}. Generally speaking, giving the AI more detailed style references leads to much higher quality results \cite{krishnan2021textstylebrush}. Even though these AI pictures look amazing, the majority of current methods still depend on standard Convolutional Neural Networks derived from BigGAN \cite{brock2019biggan}. As discussed earlier, these networks operate with restricted receptive fields. Because they focus so intensely on tiny local details, they struggle to see the bigger picture. This prevents them from grasping the global structure of the text, often resulting in a style that does not look entirely natural. Recently, the fast development of diffusion models \cite{dhariwal2021diffusion, wang2022diffusiongan, xu2023ufogen} has opened up fresh opportunities for creating handwritten text. However, some of the initial diffusion projects \cite{nikolaidou2023wordstylist, Zhu_2023_CVPR} tied their generation process to fixed identity labels, meaning they simply cannot copy a brand new, unseen handwriting style. To fix this issue, models like DiffusionPen \cite{wang2022diffusiongan} and One-DM \cite{one-dm2024} pull visual clues directly from reference pictures and mix them with the desired text to guide the generation process. While these diffusion methods are incredibly flexible, they require massive computing power and run very slowly. This makes them highly impractical for real time uses or for running on regular devices with limited processing strength.

To overcome localized CNN constraints, recent literature explores hybrid architectures combining CNNs with Transformers to jointly model global and local dependencies~\cite{bhunia2021handwriting, pippi2023handwritten, kass2022attentionhtr}. However, self-attention incurs quadratic complexity $\mathcal{O}(N^2)$~\cite{vaswani2017attention}, introducing significant computational overhead for GAN training. As a lighter alternative, MLP-based networks~\cite{melaskyriazi2021need, Tang_2022WaveMlp} achieve competitive performance with far fewer parameters, motivating FW-GAN~\cite{TONGDANGKHOA2026130175} to integrate MLPs into the generation process. Despite previous MLP-based efforts such as FW-GAN, these modules have yet to be explored in the discriminator, where CNN backbones still dominate, discarding small details due to their localized receptive fields. This gap motivates us to explore the combination of CNN and MLP modules in the discriminator, aiming to develop a handwriting framework that balances high-quality evaluation with computational efficiency. 

Furthermore, in FW-GAN~\cite{TONGDANGKHOA2026130175}, the authors achieved state-of-the-art one-shot handwriting synthesis using Wave-MLP~\cite{Tang_2022WaveMlp}. While effective, Wave-MLP is limited to small local windows, and other MLP variants such as CycleMLP~\cite{chen2023cyclemlp} and ASMLP~\cite{lian2021mlp} adopt criss-cross mixing along horizontal and vertical axes,both might be failing to model the curved and diagonal movements of handwritten images. But, the full receptive field might be inefficient for CV tasks~\citep{mu2025spiralmlp}. This insight is given by AttentionViz~\citep{yeh2023attentionvizglobalviewtransformer}, which revealed that transformer attention heads naturally exhibit spiral-like patterns. Building on this observation, SpiralMLP~\citep{mu2025spiralmlp} demonstrates that MLPs can achieve competitive performance on vision tasks by adopting carefully designed spiral offsets with lightweight complexity. Continuous handwriting also follows a non-grid structure, where stroke trajectories do not conform to rigid axes, thereby demanding customized spatial modeling. Motivated by these insights, we aim to design a specialized shape that can better capture natural strokes while keeping reasonable complexity.

CNNs~\citep{2012imagenet, he2016deep, simonyan2014very, liu2022convnet} and Transformers~\citep{vaswani2017attention, dosovitskiy2021image} are impractical for high-resolution synthesis, limited by linear aggregation or heavy cost~\citep{ma2024rewrite}. FocalNet~\citep{yang2022focal}, HorNet~\citep{rao2022hornet}, and VAN~\citep{guo2023visual} use element-wise multiplication for low-cost feature fusion but lack theoretical grounding~\citep{ma2024rewrite}. Then, Rewrite the Stars~\citep{ma2024rewrite} first showed the star operation implicitly expands features into a high-dimensional non-linear space without extra cost, and introduced StarNet as a lightweight yet powerful architecture. Moreover, the star operation has since been adopted in later works, such as RSPNet~\citep{long2026rspnet} for road surface perception, while CW-Measure-pose~\citep{xu2025automated} and RTDetrCrack~\citep{yu2025lightweight} utilize StarNet as their backbone for keypoint detection and crack detection, respectively, consistently achieving strong accuracy with minimal computational cost. These vision tasks share a common challenge with handwriting synthesis: Capturing fine-grained details while maintaining modest computation. Building on these motivations, we adopt this special operation in our generation process, offering a potential improvement in synthesis quality.

Aside from the structural limits we just discussed, most current methods completely ignore information found in the frequency domain. A concept known as the F Principle \cite{Zhi_2020FPrinciple} explains that when neural networks tend to learn low-frequency (smooth) components of the target function before capturing high-frequency (detailed) components during training. This phenomenon has been theoretically supported in diverse settings, including infinite data regimes~\cite{luo2019theory}, wide neural networks under the Neural Tangent Kernel (NTK) framework~\cite{jacot2018ntk}, and finite-sample scenarios~\cite{luo2020theory, zhang2019explicitizing, bordelon2020spectrum, cao2019towards, basri2019convergence, yang2019fine}. Furthermore, E et al.~\cite{e2019frequency} show that this behavior can naturally emerge from the integral formulation of network optimization. Inspired by these findings, recent works have incorporated frequency-domain constraints into generative models. In handwriting synthesis, FW-GAN~\citep{TONGDANGKHOA2026130175} adopts the Frequency Distribution Loss (FDL)~\citep{Ni_2024FDL} to match the spectral distributions of real and synthetic samples. This motivates us to integrate similar frequency-aware components into our pipeline to further enhance generation quality.

Furthermore, handwriting synthesis shares common challenges with document enhancement, as both tasks involve generating text-laden images where high-frequency details are often ignored or lost. In document enhancement, regression-based methods such as DocEnTr~\citep{souibgui2022docentr} and DE-GAN~\citep{souibgui2020gan} optimized for pixel-level losses tend to suffer from significant loss of high-frequency information, leading to distorted and blurred text edges~\citep{yang2023docdiff}. To address this, DocDiff~\citep{yang2023docdiff} demonstrates that explicit edge supervision can effectively sharpen text edges. In handwriting synthesis, however, existing approaches such as FW-GAN~\citep{TONGDANGKHOA2026130175}, despite incorporating a high-frequency discriminator, do not clearly improve edge sharpness or concentrate on edge supervision like DocDiff. Driven by these observations, we propose an edge-guided components, offering a potential direction for improving handwriting synthesis performance.

Drawing from those, we introduce SpiS-GAN, a hierarchical generative framework designed for high-quality, one-shot handwriting synthesis. We replace traditional convolutional blocks with Star-Spiral Blocks. These blocks incorporate our Modulated Elliptical SpiralFc to dynamically track ink flow along connected trajectories, and utilize the special element-wise multiplication to implicitly expand the model's feature space. This combination allows for highly detailed, variable-length text generation while keeping computational costs acceptable. To strictly enforce shape integrity, we introduce a edge-aware loss, which penalizes shape distortions and prevents the model from producing blurred letter boundaries. Furthermore, we deploy a dual discriminator architecture where a standard discriminator enforces visual realism, while a dulated Discriminator jointly evaluates multi-domain features that capture broken connections and shape anomalies. We adopt the HiGAN~\cite{gan2021higan} framework, augmented with a Frequency Distribution Loss (FDL)~\cite{Ni_2024FDL}, as our practical baseline for its stable training behavior and compatibility with BigGAN-style generators. However, our proposed adversarial modifications are independent and can be easily integrated into alternative generative models. An overview of the proposed framework is illustrated in Figure~\ref{fig:arch}.

\section{Proposed Approach}\label{sec:proposal}
\subsection{Problem Formulation}\label{subsec:formulation}

Our core objective in one-shot handwriting synthesis is to produce realistic text images using just a single reference sample. Let $\mathbf{x}$ represent a handwritten word image from a particular writer $\mathbf{w}$, which inherently carries a unique calligraphic style $\mathbf{z}$. We want to generate new handwritten images that faithfully preserve this writer's visual identity. We define a set of textual queries $A = \{a_k\}_{k=1}^Q$, where $Q$ denotes the total number of queries, and each $a_k$ is a target word of arbitrary length selected from a general character vocabulary. More precisely, each query word is represented as a discrete character sequence $a_k = [a_{k,1}, \ldots, a_{k,L_k}]$, with $L_k$ indicating the character length of the $k$-th word. To produce the final synthetic image $\hat{\mathbf{x}}_k$, we train a generator $G$ that effectively combines the textual content with the extracted style. The generator acts as a mapping function that projects the rigid character sequence $a_k$ and the dynamic style vector $\mathbf{z}$ into the visual domain:
\begin{equation}
	\hat{\mathbf{x}}_k = G(a_k, \mathbf{z}).
\end{equation}
The style vector $\mathbf{z}$ can be obtained in two ways. It can either be randomly sampled from a standard normal distribution $\mathcal{N}(0, \mathbf{I})$, enabling diverse style variations, or it can be directly extracted from the reference image $\mathbf{x}$ using a style encoder $E$, such that $\mathbf{z} = E(\mathbf{x})$. This dual approach offers flexibility: sampling allows us to generate novel stylistic variations during inference, while extraction ensures that the generator can reproduce the exact handwriting style of the target writer. By supporting both modes, our framework accommodates a wide range of use cases, from creative style exploration to precise style replication.

	\begin{figure}[!t]
	\footnotesize
	\centering
	\includegraphics[width=1\linewidth]{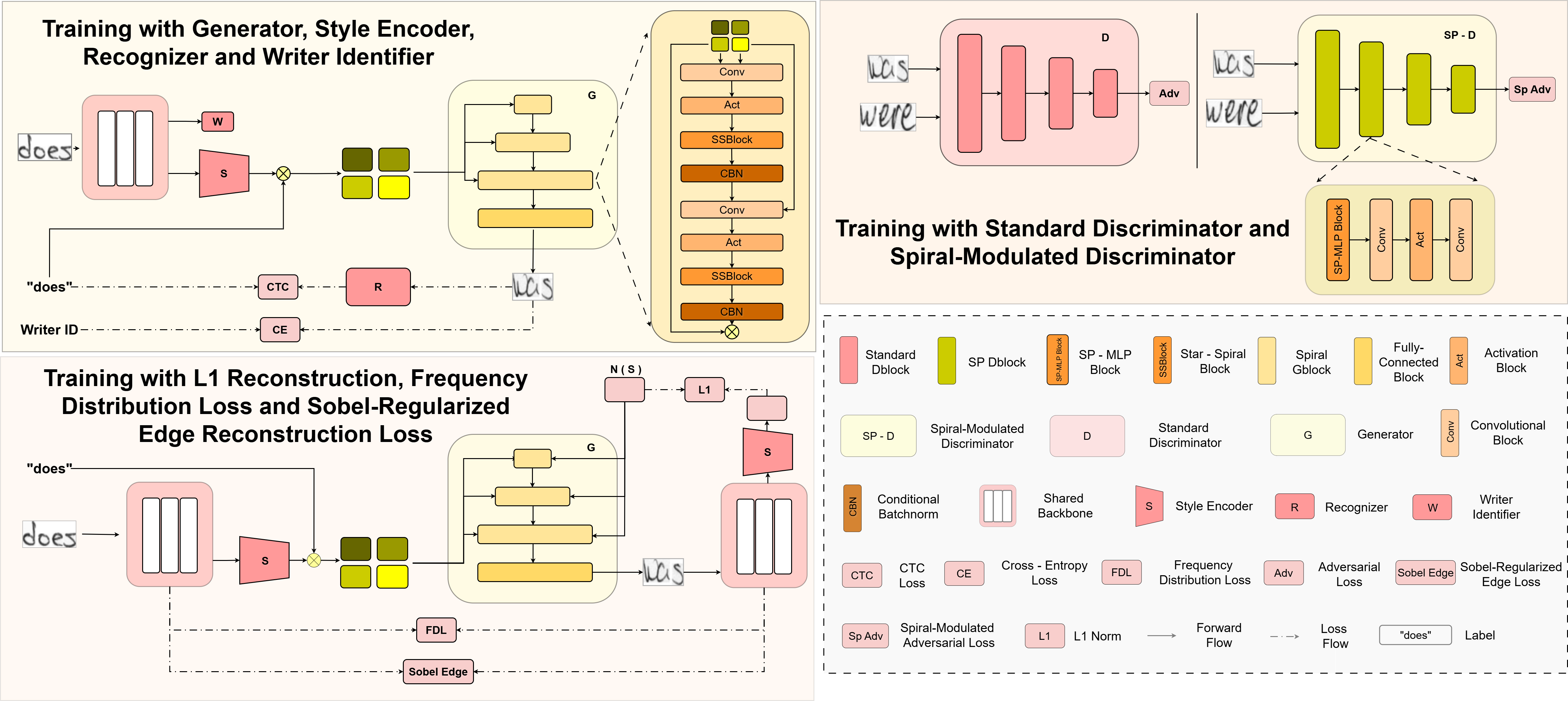}
	\caption{Overview of the SpiS-GAN architecture. The input text is converted to one-hot and combined via element-wise multiplication with a style feature, either sampled from a Gaussian or encoded from a reference image. The style encoder splits the encoded style into segments that are progressively injected into the generator, which upsamples and refines features to synthesize handwritten text. At inference, only the style encoder and generator are needed.}
	\label{fig:arch}
	\footnotesize
\end{figure}

\subsection{Overall Architecture}
\label{sec:architecture}

Our model features a hierarchical, style-conditioned generator composed of Star-Spiral Blocks (SSB), which combine the StarNet-style self-multiplication
and Modulated Elliptical SpiralFC (MESpiralFC) to effectively model shape, spatial, and style variations in flowing handwriting, supporting variable-length text generation. The generator is conditioned on character-level content embeddings and writer-specific style features, enabling personalized synthesis. To improve realism, we employ a dual-discriminator architecture consisting of a standard discriminator and a novel discriminator adopting a MLP-based design to identify geometric anomalies, that are not fully captured by spatial analysis. The framework also includes a recognizer for content supervision, a style encoder for extracting writer-specific characteristics, and a writer identifier to maintain style consistency during training. All components are jointly optimized in a unified min-max framework with adversarial objectives and auxiliary losses, including Frequency Distribution Loss (FDL) for global alignment and Sobel-Regularized Edge Reconstruction Loss (SELoss) for local frequency refinement, ensuring accurate handwriting synthesis consistent with the target writing style. 

\subsubsection{Modulated Elliptical SpiralFC}
\label{sec:spiral}

\begin{figure}[!t]
	\centering
	\includegraphics[width=0.9\textwidth]{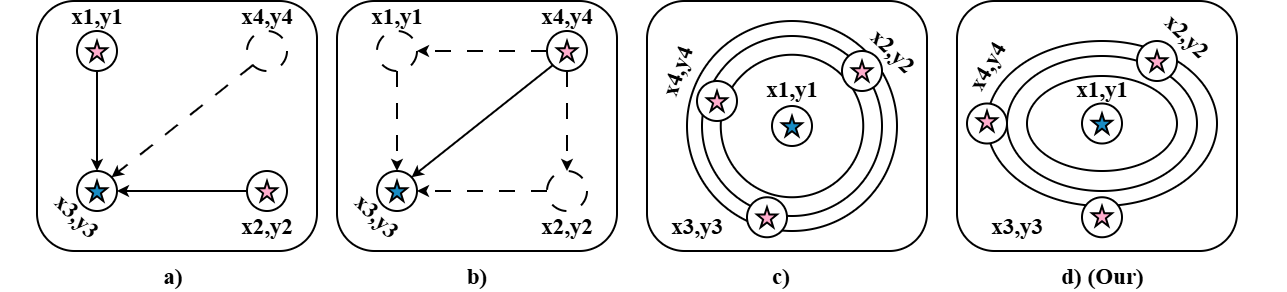}
	\caption{(a)–(b) Grid-based; (c) Original SpiralFC; (d) Elliptical SpiralFC (ours).}
	\label{fig:spiral_comparison}
\end{figure}

We build upon SpiralMLP~\citep{mu2025spiralmlp}, whose SpiralFC (Fully-Connected) layer is designed as a deformable convolution-style layer with predefined spiral offsets, achieving low computation. Given a feature map \(X\in\mathbb{R}^{H\times W\times C_{\text{in}}}\) with input channel dimension \(C_{\text{in}}\), Spiral FC produces an output with \(C_{\text{out}}\) channels.
\begin{equation}
	\text{Spiral FC}_{i,j,:}
	(X) = \sum^{C_{\text{in}}}_{c=0}X_{i+\phi_i(c),j+\phi_j(c),c}W^{\text{spiral}}_{c,:}+b^{\text{spiral}},
	\label{eq:spiralfc}
\end{equation}
where \(W^{\text{spiral}}\in\mathbb{R}^{C_{\text{in}}\times C_{\text{out}}}\) and \(b^{\text{spiral}}\in\mathbb{R}^{C_{\text{out}}}\) are the trainable matrix and bias, \(\text{Spiral FC}_{i,j,:}(\cdot)\) denotes the output at position \((i,j,:)\), and \(\phi_i(c)\) and \(\phi_j(c)\) are the offset functions along the \(H\) and \(W\) axes, respectively, with \(c\) denoting the channel index (\(0 \le c < C_{\text{in}}\)). Since each output location samples only one spatial position from each input channel, Spiral FC requires only $C_{\text{in}}$ sampled values. The original Spiral FC defines the offsets as
\begin{align}
	\phi_i(c) = A(c)\cos{(\frac{c\times2\pi}{T})},
	\label{eq:orig_offset_i}\\
	\phi_j(c) = A(c)\sin{(\frac{c\times2\pi}{T})},
	\label{eq:orig_offset_j}
\end{align}
where $T$ is the constant period controlling the spiral rotation, and $A(c)$ is the amplitude function (see Eq.~6 in~\citep{mu2025spiralmlp}). Although efficient, this circular shape(Figure~\ref{fig:spiral_comparison}c) is not well suited to handwriting because it samples equally in all directions, ignores the natural left-to-right writing flow, and uses a fixed sampling radius across network stages. In contrast, existing criss-cross approaches (Figure~\ref{fig:spiral_comparison}a–b) restrict sampling to horizontal and vertical axes, failing to capture diagonal and curved movements.

To address these limitations, we propose the Modulated Elliptical SpiralFC, a variant of Spiral FC that applies a channel-wise modulation to its trajectory and replaces the circular path with an elliptical shape (Figure~\ref{fig:spiral_comparison}d). We split the channels into \(k\) groups following SpiralMLP~\citep{mu2025spiralmlp}, but set the width of each group to \(W = C_{\text{in}} / k\). Unlike the original SpiralFC, where the global amplitude \(A(c)\) and global channel index \(c\) define a single spiral, our design applies the same scaling function \(\mathcal{R}(i)\) independently within each group, where the local index \(i\) resets from \(0\) to \(W-1\). This yields multiple identical elliptical loops instead of just one. The offsets are defined as:
\begin{align}
	\phi_i(c) = \mathcal{R}(i)R_y\sin{(\frac{i\times2\pi}{T})},
	\label{eq:mespiralfc_offset_i}\\
	\phi_j(c) = \mathcal{R}(i)R_x\cos{(\frac{i\times2\pi}{T})},
	\label{eq:mespiralfc_offset_j}
\end{align}
where \(T\) follows the original formulation, and \(R_x, R_y\) control the horizontal and vertical extents. Setting \(R_x > R_y\) stretches the pattern horizontally, thereby creating elliptical patterns to better match the natural writing direction of human handwriting. Moreover, the modulation function \(\mathcal{R}(i)\) follows a triangular pattern:
\begin{equation}
	\mathcal{R}(i)=
	\begin{cases}
		\dfrac{i}{W/2}, & 0\le i\le W/2,\\[4pt],
		\dfrac{W-i}{W/2}, & W/2<i\le W.
	\end{cases}
\end{equation}
This expands and contracts the sampling scope within each partition, allowing each loop to capture features at varying distances from the center. Since \(R_x\) and \(R_y\) are defined relative to \(C_{\text{in}}\), the sampling extent automatically adapts across network stages, removing the fixed constraint of SpiralFC while preserving linear computational complexity~\citep{mu2025spiralmlp}.

\begin{figure}[!t]
	\footnotesize
	\centering
	\includegraphics[width=0.8\linewidth]{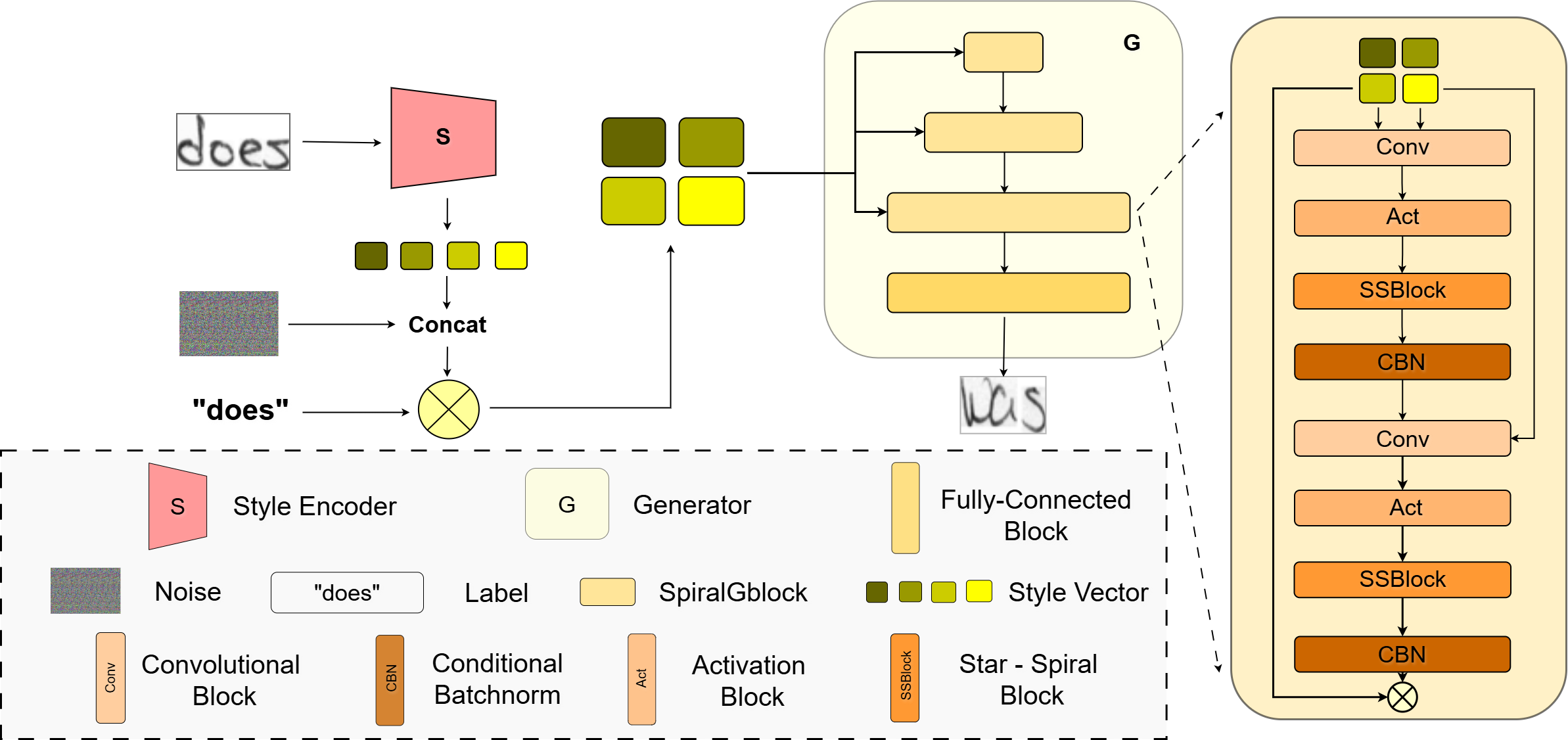}
	\caption{Overview of the SpiS-GAN hierarchical generator featuring Star-Spiral Blocks}
	\label{fig:generator}
	\footnotesize
\end{figure}

\subsubsection{Hierarchical Generator and the Star-Spiral Blocks}
\label{sec:generator}
Our handwriting synthesis framework employs a hierarchical, style-conditioned generator that takes as input both a latent style vector $\mathbf{z} \in \mathbb{R}^d$ and a character sequence $\mathbf{y} \in \{0,1\}^{n \times L}$, where $n$ is the vocabulary size and $L$ is the sequence length. During training, we randomly draw $\mathbf{z}$ from a Gaussian prior distribution $\mathcal{N}(0,1)$. The style vector is divided into several segments. The first segment is combined with the embedded character sequence via element-wise multiplication, after which it is projected and reshaped to create the initial latent feature map \(\mathbf{F}_0 \in \mathbb{R}^{C_0 \times H_0 \times W_0}\), where \(C_0\) denotes the starting number of channels and \(H_0, W_0\) represent the spatial dimensions of the feature map. The remaining segments are progressively fed into the SpiralGblocks, which serve as replacements for the standard BigGAN blocks in our generator, through Conditional Batch Normalization (CBN). This mechanism allows each stage to adjust its feature representations according to the desired writing style. Each SpiralGblock consists of two Star-Spiral Blocks separated by an upsampling layer. Through this process of gradual upsampling and iterative feature refinement, the generator ultimately produces handwriting images that accurately preserve both the given text content and the intended calligraphic style.

\begin{figure}[!t]
	\centering
	\includegraphics[width=0.8\textwidth]{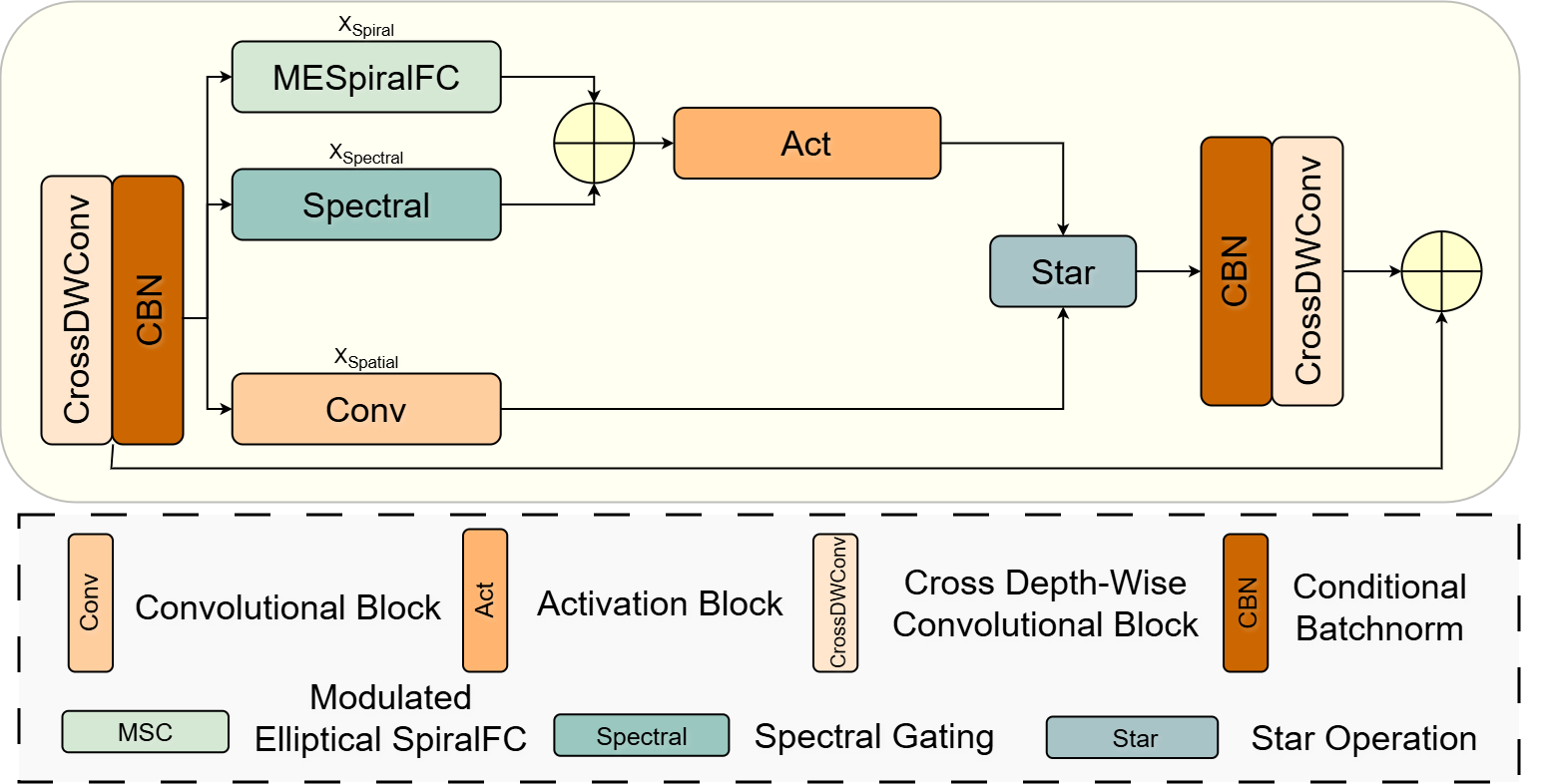}
	\caption{Detailed architecture of Star-Spiral Block}
	\label{fig:ssb}
\end{figure}

Handwriting generation requires modeling tightly coupled interactions between character structure, stroke trajectories, and writing style. To capture such complex dependencies efficiently, we adopt the Star self-multiplication \citep{ma2024rewrite}. For a feature vector $x \in \mathbb{R}^{d+1}$ at each spatial location (with $d$ denoting the input channel dimension) and learned weights $w_1, w_2 \in \mathbb{R}^{d+1}$, the operation is defined as (see Eq.~(1) in \citep{ma2024rewrite}):
\begin{align}
	w_1^\mathrm{T}x * w_2^\mathrm{T}x
	&=
	\left(
	\sum_{i=1}^{d+1} w_1^i x^i
	\right)
	*
	\left(
	\sum_{j=1}^{d+1} w_2^j x^j
	\right)
	\\
	&=
	\sum_{i=1}^{d+1}
	\sum_{j=1}^{d+1}
	w_1^i w_2^j x^i x^j,
	\label{eq:star_fuse}
\end{align}
where \(i\) and \(j\) index the channel dimension. This expands into \(\frac{(d+1)(d+2)}{2}\) pairwise interaction terms (see Eq.~(4) in \citep{ma2024rewrite}). When stacking \(l\) such layers, the interaction complexity grows exponentially with depth (see Eq.~(10) in \citep{ma2024rewrite}):
\begin{equation}
	O_l
	=
	W_{l,1}^{T}O_{l-1}
	\ast
	W_{l,2}^{T}O_{l-1},
	\label{eq:star_exponential}
\end{equation}
where \(O_l\) is the output feature map at layer \(l\), \(O_{l-1}\) is the output from the previous layer, and \(W_{l,1}, W_{l,2}\) are the corresponding weight matrices. The core of StarNet lies in its star operation mechanism, which dynamically aggregates features from multiple branches through element-wise multiplication, enabling efficient fusion and low-dimensional computation. Unlike the serial structure of traditional CNNs, this operation implicitly maps features into a high-dimensional non-linear space without increasing channel width, preserving efficiency while enabling rich higher-order feature modeling.

\begin{figure}[!t]
	\centering
	\includegraphics[height=0.5\textwidth]{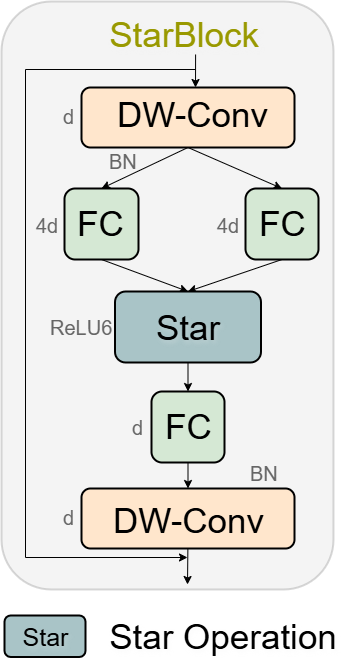}
	\caption{Architecture of the original StarBlock from StarNet~\citep{ma2024rewrite}.}
	\label{fig:starb}
\end{figure}
We build the Star-Spiral Block (SSB) (Figure~\ref{fig:ssb}) upon the original StarBlock design (Figure~\ref{fig:starb}). Although StarBlock was originally introduced as a proof-of-concept architecture, its effectiveness as a lightweight backbone has been validated by recent methods in diverse tasks (detailed in Section~\ref{sec:offline_handwriting_generation}), motivating our adoption of it as the foundation for handwriting generation. We adapt it through three key modifications: First, we replace the standard \(7 \times 7\) depthwise convolutions at both ends of the block with split filters (\(1 \times 7\) and \(7 \times 1\)), referred to as cross-depthwise convolution in Figure~\ref{fig:ssb}, decomposing spatial processing into horizontal and vertical components to better capture stroke directions while reducing parameters. Second, we redesign the two original parallel FC branches: one is enhanced with the MESpiralFC (Section~\ref{sec:spiral}), which captures rich spatial context to produce \(X^{\text{spiral}}\), and a Spectral Gating module that applies a real FFT, scales frequency components with learnable complex weights, and transforms back to the spatial domain to produce \(X^{\text{spectral}}\), enhancing informative frequency components. The other branch remains a pointwise convolution to preserve structural layout, producing \(X^{\text{spatial}}\).  Following the original paper and our own experiments, we apply an activation \(\sigma\) to only one branch before element-wise multiplication, acting as a gating mechanism:
\begin{equation} \label{eq:star_fused}
	X^{\text{fused}} = \sigma(X^{\text{spiral}} + X^{\text{spectral}}) \ast X^{\text{spatial}}.
\end{equation}
This operation enables SSB to learn complementary spatial, spectral, and structural dependencies, resulting in more consistent style preservation across generated samples while preserving channel efficiency. The first novel block operates on low-resolution features to capture global structure, while the second refines the upsampled features for detailed handwriting generation. By jointly modeling global context and local style, the generator produces coherent, high-quality handwriting that faithfully maintaining both the input text and the writer's style.

\begin{figure*}[!t]
	\footnotesize
	\centering
	\includegraphics[width=1\linewidth]{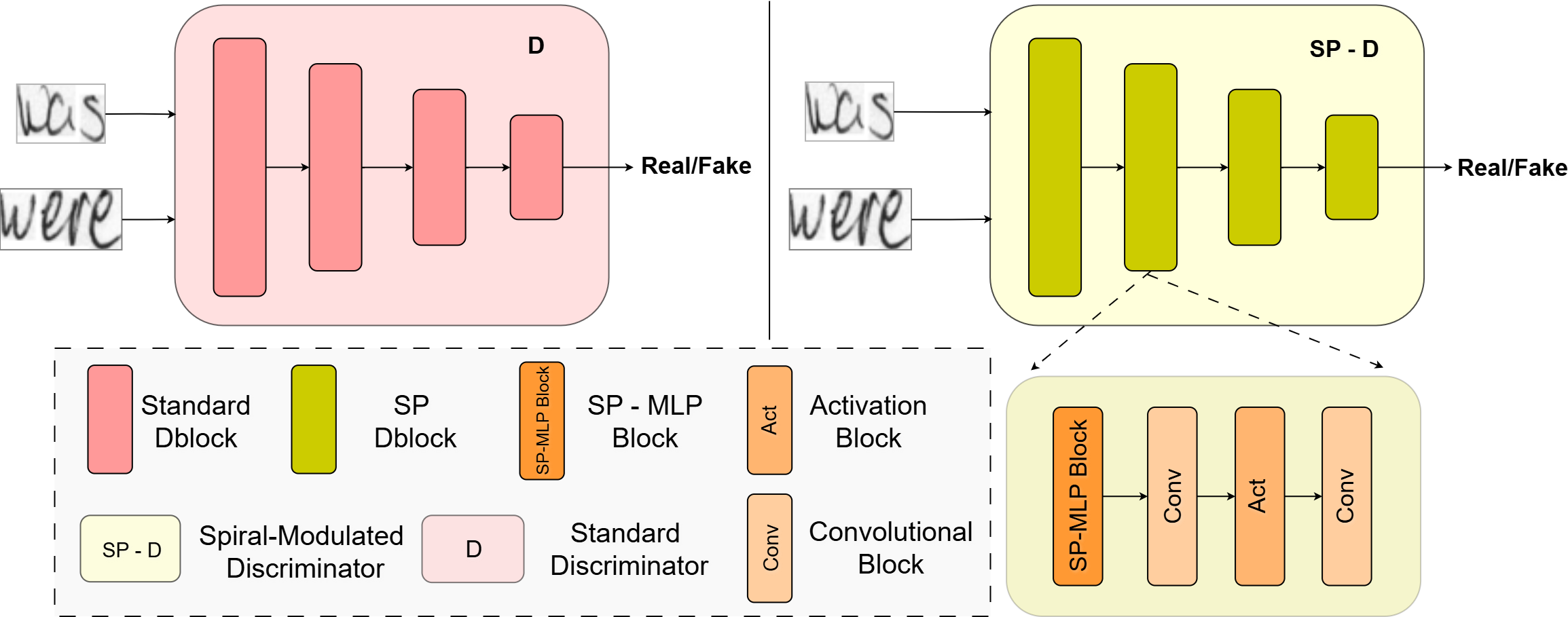}
	\caption{Illustration of the Dual Discriminator Architecture.}
	\label{fig:disc-arch}
	\footnotesize
\end{figure*}

\subsubsection{Dual Discriminator Architecture}
\label{subsubsec:discriminator}

To strengthen adversarial supervision, we employ a dual-discriminator framework consisting of a conventional spatial discriminator $D$ and the proposed discriminator, as illustrated in Figure~\ref{fig:disc-arch}.

\paragraph{\textbf{Standard Spatial Discriminator}}
The spatial discriminator $D$ functions as a traditional adversarial network that processes handwriting images directly in the spatial domain. Its architecture is composed of a series of convolutional blocks, each made up of a convolution layer followed by spectral normalization, batch normalization, and a ReLU activation function. As the network goes deeper, the spatial resolution gradually decreases while the number of feature channels increases, enabling the model to extract features at multiple levels, from fine stroke textures and individual character shapes to broader structural patterns. The final feature map is then condensed through global average pooling and passed through a linear layer to produce the realism score $D(x)$, which indicates how authentic the input handwriting appears.

\begin{figure}[!t]
	\centering
	\includegraphics[width=0.8\textwidth]{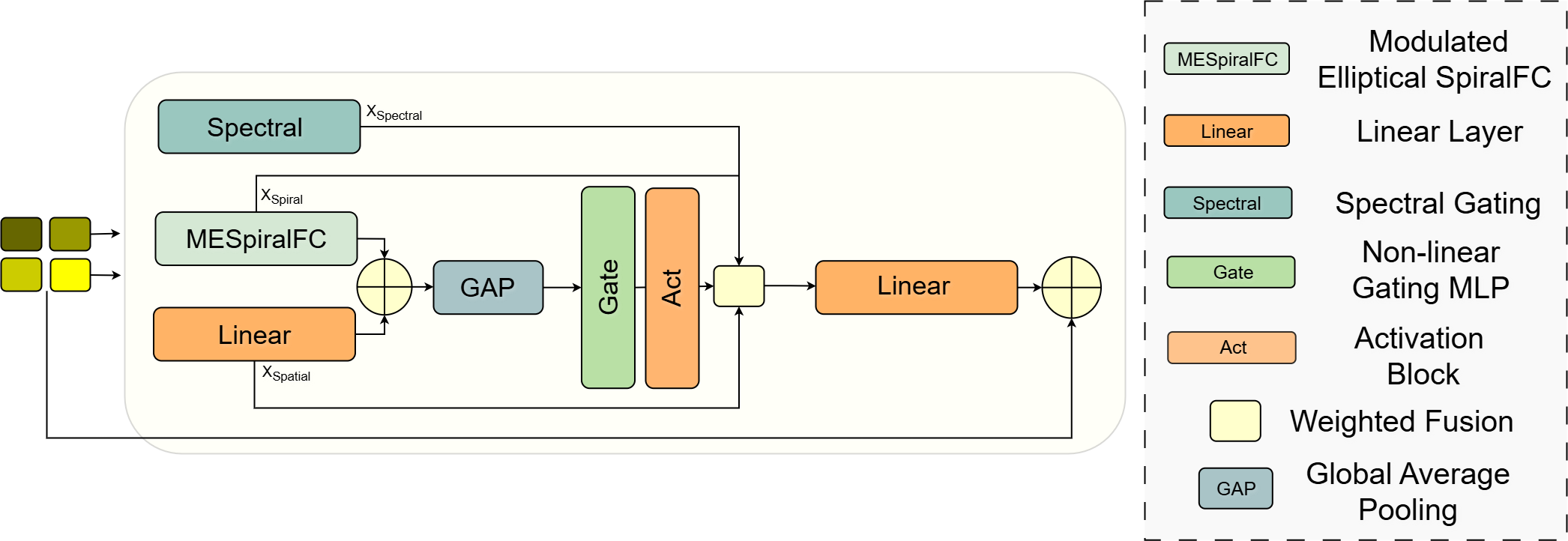}
	\caption{Detailed architecture of SP-MLP Block.}
	\label{fig:SP-MLP}
\end{figure}

\paragraph{\textbf{Spiral-Modulated Discriminator}}

Conventional CNN-based discriminators suffer from localized receptive fields and aggressive downsampling, which together discard fine-grained details and cause locally consistent strokes to appear globally broken or disconnected. To address this, we replace the standard convolutional block in the discriminator with our proposed SPDBlock, which applies a customized SpiralMLP (SP-MLP Block) before spatial pooling (see Figure~\ref{fig:SP-MLP}). Our approach employs a refined feature extraction strategy that captures sufficient spatial context without requiring full receptive field coverage.

We decompose feature evaluation into three pathways, each producing a feature map of size \(\mathbb{R}^{H \times W \times C_{out}}\) (with \(C_{\text{out}}\) as defined in Section~\ref{sec:spiral}). A spatial linear layer produces \(X^{\text{spatial}}\) (similar to Self-Spiral FC~\citep{mu2025spiralmlp}) to detect pixel-level anomalies. MESpiralFC generates \(X^{\text{spiral}}\) (a variant of Cross-Spiral FC) to model stroke integrity; interruptions along its curved motion reveal flaws missed by grid-based modules. Spectral Gating~\citep{rao2021global} produces \(X^{\text{spectral}}\) via 2D FFT to identify frequency anomalies, preventing the generator from masking blurriness with artificial high-frequency noise. In the original SpiralMLP~\citep{mu2025spiralmlp}, Spiral Mixing consists of Self-Spiral FC, Cross-Spiral FC, and a Merge Head that fuses the two branches via SoftMax-weighted summation. To extend this design to our three-pathway architecture, we redesign the Merge Head of SpiralMLP (Eq.~7-8 in~\citep{mu2025spiralmlp}) as detailed in Figure~\ref{fig:SP-MLP}. In the baseline, features from parallel Spiral FC branches (\(X^{\text{self}}, X^{\text{cross}} \in \mathbb{R}^{H \times W \times C_{out}}\)) are fused using a mixing weight matrix \(a \in \mathbb{R}^{2 \times C_{out}}\) derived from global average pooling and SoftMax:
\begin{equation}
	a=\sigma(W^{\text{merge}}\times[\frac{1}{HW}\sum^{HW}_{i=0}\mathcal{F}(X^{\text{self}}+X^{\text{cross}})_{i,:}]),\label{eq:9}
\end{equation}
where $\mathcal{F}$ flattens spatial dimensions, and \(W^{\text{merge}}\) is a learnable projection matrix that maps the pooled context to the mixing weight space. Here, \(a\) is the mixing weight matrix from the SoftMax-based Merge Head, whose components \(a_{1,:}\) and \(a_{2,:}\) correspond to the two branches \(\mathbf{X}_{self}\) and \(\mathbf{X}_{cross}\), respectively. To extend this design to our three-pathway architecture, we replace \(X^{\text{self}}\) with \(X^{\text{spatial}}\), \(X^{\text{cross}}\) with \(X^{\text{spiral}}\), and add \(X^{\text{spectral}}\) as a third branch. We extend the two-way mixing to three independent gates \(a_1, a_2, a_3 \in [0,1]^{C_{out}}\). The gate context is computed from the same two branches as the baseline, excluding the added one to prevent frequency bias. Each gate scales its corresponding branch via element-wise multiplication. At a specific spatial position \((i,j)\), the fused representation is:
\begin{equation}
	X^{\text{fused}}_{i,j,:}
	=
	\left(
	a_{1,:}\odot X^{\text{spiral}}_{i,j,:}
	+
	a_{2,:}\odot X^{\text{spatial}}_{i,j,:}
	+
	a_{3,:}\odot X^{\text{spectral}}_{i,j,:}
	\right).
	\label{eq:fused}
\end{equation}
Across the entire feature map, the gates are broadcast to influence all spatial positions. This independent gating enables pixel-level defects to be modeled in parallel, providing a more comprehensive evaluation signal. The gated combination is then passed through a linear projection and subsequently processed through the standard convolutional backbone, yielding a realism score that assesses both global stroke consistency and fine-scale textural realism in the generated cursive samples (see Figure~\ref{fig:SP-MLP}).

\subsubsection{Recognizer}
\label{subsec:text_recognizer}

The recognition component, denoted as $R$, is responsible for extracting the textual sequence $y$ directly from the provided handwritten images. This network is optimized exclusively utilizing authentic handwritten datasets equipped with manual annotations, intentionally excluding any artificially generated samples from its training phase. Throughout the synthesis stage, $R$ functions as an independent verification mechanism, evaluating the output of the generator $G$ to enforce strict character precision. Consequently, this architecture ensures that the synthesized script not only preserves visual coherence but also aligns flawlessly with the designated target string.

\subsubsection{Style Encoder and Writer Identifier}
\label{subsec:style_writer_modules}

The style encoding module $E$ captures visual aesthetics from images of written words and compresses them into a latent vector $\mathbf{s}$ of constant dimensions, serving as a conditional input for the generator $G$. Simultaneously, the writer identification component $W$ determines the specific author of the provided script, delivering direct stylistic supervision throughout the optimization phase. These two components operate under a unified primary goal, which involves isolating features associated with penmanship style while completely ignoring the underlying text. This mutual objective justifies the implementation of a common base architecture, utilizing the exact same framework presented in \citep{TONGDANGKHOA2026130175}. This shared foundation evaluates incoming visual data via layered convolutional operations integrated with residual pathways. By utilizing a unified feature extraction process, the framework significantly lowers computational demands while enabling both operations to leverage identical visual understandings.

Following this shared foundational network, every component branches into an independent prediction layer. Specifically, the encoding module applies a variational strategy to map the statistical distribution of handwriting aesthetics, whereas the identification module conducts a categorical classification to recognize individual authors. Furthermore, both systems incorporate specialized mechanisms capable of processing input sequences of irregular lengths. By adopting a joint learning strategy, this structural design exploits the synergy between style derivation and author classification. These two complementary objectives actively support one another, ultimately yielding highly comprehensive and resilient visual representations for the entire system.

\subsection{Objective Functions}

Before the optimization phase, our architecture necessitates a collection of data offering multiple layers of supervision. This required information includes visual samples of penmanship ($\mathbf{X}$), their corresponding textual annotations ($\mathbf{Y}$), and the specific author categories ($\mathbf{W}$). Even though these labeled instances act as the foundational elements for optimization, our generative pipeline is required to extrapolate beyond this limited scope. This extrapolation is especially critical when the system needs to render entirely new sequences or words that out-of-vocabulary (OOV). To resolve this challenge, we extract textual sequences from a significantly larger linguistic collection. This strategy permits the architecture to process random strings of characters while learning, which is achieved by selecting the target string $\tilde{y}$ directly from an expansive text corpus $\mathcal{C}$. The comprehensive learning methodology, encompassing structural decisions and the exact formulation of error functions, is visually outlined in Figure~\ref{fig:arch} and discussed in detail in the following paragraphs.

\subsubsection{Adversarial Loss}
\label{subsec:adversarial_loss}

Our framework implements a classic Generative Adversarial Network architecture. In this arrangement, the discrimination module $D$ is trained to distinguish authentic penmanship specimens from artificial images created by the generative component $G$. This competitive dynamic forces $G$ to synthesize script that exhibits a significantly higher degree of visual realism. To ensure a steady optimization process and facilitate robust learning, we incorporate the popular hinge loss mathematical structure \cite{lim2017geometric}, which is expressed mathematically as follows:
\begin{equation}
	\mathcal{L}_{adv} = \mathbb{E}_{x \sim \mathbf{X}} \left[ \max(0, 1 - D(x)) \right] + \mathbb{E}_{\tilde{y} \sim \mathcal{C},\ z} \left[ \max(0, 1 + D(G(\tilde{y}, z))) \right].
\end{equation}
In this context, the stylistic vector $\mathbf{z}$ is acquired through one of two methods. The first method involves sampling from a standard normal distribution \(\mathcal{N}(0,1)\). The second method requires processing a target reference picture \(x\) using the encoding network, resulting in $\mathbf{z} = E(\mathbf{x})$. Readers should observe that this competitive penalty serves solely to elevate the aesthetic quality of the artificial outputs. This specific metric completely ignores the accuracy of the textual sequence and the precise replication of the author aesthetics. Those particular requirements are handled by distinct optimization functions integrated elsewhere in our design.

\subsubsection{Text Recognition Loss}

To ensure the synthetic script faithfully retains the desired character sequence, we integrate a pre-trained recognition network $R$ to guide the generative model $G$. Although the adversarial objective promotes visual authenticity, it fails to guarantee that the artificial images contain accurate linguistic information. The identification module $R$ bridges this gap by directly aligning the visual features of the generated image with the specified textual targets. In its initial phase, this network undergoes supervised learning using authentic handwritten specimens $x \in \mathbf{X}$ paired with their corresponding transcriptions $y \in \mathbf{Y}$. We utilize the Connectionist Temporal Classification (CTC) framework \cite{Graves2006ctc} to manage unsegmented sequence learning. This relationship is mathematically defined as follows:
\begin{equation}
	\mathcal{L}_{R}^{D} = \mathbb{E}_{x, y} \left[ \mathcal{L}_{\text{CTC}}(R(x), y) \right],
\end{equation}
In this expression, $R(x)$ represents the series of estimated character probabilities for a given picture $x$, while $y$ signifies the ground-truth transcript.

After $R$ reaches complete convergence, its weights are permanently frozen. From that moment, it acts as a perceptual guide to preserve textual fidelity within the creations produced by the generator. During practical execution, when given a randomly chosen string $\tilde{y} \in \mathcal{C}$ and a style vector $z$, the network $G$ learns to fabricate a visual representation $G(\tilde{y}, z)$. The optimization requires that when $R$ analyzes this new picture, it successfully decodes the original string $\tilde{y}$. This requirement introduces the generator-side recognition loss:
\begin{equation}
	\mathcal{L}^{G}_{R} = \mathbb{E}_{\tilde{y}, z} \left[ \mathcal{L}_{\text{CTC}}(R(G(\tilde{y}, z)), \tilde{y}) \right].
\end{equation}
The resulting error signal is propagated backward through the layers of $G$, continuously improving its capacity to construct syntactically accurate penmanship. Ultimately, this methodological choice guarantees that the artificial creations are visually convincing while simultaneously carrying the exact requested textual information.

\subsubsection{Writer Identification Loss}

While the visual characteristics of penmanship exhibit significant diversity, they tend to remain remarkably stable for any specific author. This inherent stability enables our framework to utilize author identity as an effective proxy for explicit style conditioning. Because detailed annotations for fine-grained stylistic elements (such as stroke thickness, inclination, or curve geometry) are unavailable, we introduce a writer classification module $W$ to capture these latent stylistic patterns. This classifier undergoes training to recognize the author from authentic handwritten specimens $x \in \mathbf{X}$ by leveraging their associated identity labels $w \in \mathbf{W}$. This optimization is achieved via a conventional cross-entropy loss formulation:
\begin{equation}
	\mathcal{L}_{W}^{D} = \mathbb{E}_{x, w} \left[ - \log p(w \mid W(x)) \right],
\end{equation}
This objective compels $W$ to isolate features that effectively differentiate between authors, thereby implicitly encoding the fundamental style information.

Following its convergence, the parameters of $W$ are permanently frozen, allowing it to function as an evaluation mechanism that guides the generator $G$ in preserving stylistic uniformity. When provided with a target image $x$, our system initially derives its style representation $z = E(x)$ before synthesizing a novel image $G(\tilde{y}, z)$ containing a randomly selected text string $\tilde{y} \in \mathcal{C}$. The framework subsequently mandates that this artificial sample be categorized by $W$ as originating from the exact same author as the target reference:
\begin{equation}
	\mathcal{L}^{G}_{W} = \mathbb{E}_{x, w, \tilde{y}} \left[ - \log p(w \mid W(G(\tilde{y}, E(x)))) \right].
\end{equation}

This regulatory process guarantees that the generative model faithfully adheres to the stylistic identity captured from the reference input, even while rendering completely novel sequences of text. It is important to acknowledge that the identification network $W$ is exclusively optimized on the primary training distribution and might struggle to generalize to individuals outside this dataset (for example, unseen authors present within the testing phase).

\subsubsection{Style Reconstruction Loss}

To ensure the generative module adequately incorporates the target stylistic attributes throughout the synthesis phase, we implement a structural penalty modeled after \cite{chen2016infogan}. This constraint enforces a strict alignment between the original style representation and the features extracted from the synthesized output. Establishing this relationship is essential for developing a bidirectional mapping that links the latent stylistic dimensions with the final visual characteristics of the penned text. Specifically, when provided with a stylistic variable $z \sim \mathcal{N}(0,1)$ alongside a random textual string $\tilde{y} \in \mathcal{C}$, the generative network $G$ constructs a corresponding visual representation $G(\tilde{y}, z)$. Subsequently, this artificial image is processed by the style encoder $E$, with the strict requirement that the newly extracted style vector must closely approximate the initial input:
\begin{equation}
	\mathcal{L}_{style} = \mathbb{E}_{\tilde{y}, z} \left[ \left\| z - E(G(\tilde{y}, z)) \right\|_1 \right].
\end{equation}

By incorporating this self-consistency objective, the architecture is forced to actively and meaningfully exploit the stylistic variable $z$ throughout the generative process. Furthermore, this approach promotes visual variance within the synthesized results and effectively prevents the generative network from collapsing into a limited spectrum of writing styles.

\subsubsection{KL-Divergence Loss}

To facilitate the random extraction of styles during the evaluation phase, we constrain the optimized stylistic latent space to adhere to a predetermined prior distribution. This objective is achieved by applying a Kullback-Leibler (KL) divergence constraint, which penalizes the discrepancy between the distribution of the extracted style representations and a standard normal distribution. Specifically, for any authentic handwritten specimen $x \in \mathbf{X}$, we calculate the KL-divergence metric comparing the derived posterior probability $E(x)$ against the baseline prior $\mathcal{N}(0,1)$:
\begin{equation}
	\mathcal{L}_{kl} = \mathbb{E}_{x} \left[ D_{KL}(E(x) \, \| \, \mathcal{N}(0, 1)) \right].
\end{equation}

The inclusion of this regularization metric forces the latent domain to maintain continuity and smoothness, thereby ensuring it remains highly sampleable throughout the testing procedure. This mathematical penalty serves as a fundamental element within variational architectures and has consistently demonstrated its efficacy across various style transfer methodologies \cite{zhu2017multimodal, lee2018diverse}.

\subsubsection{Frequency Distribution Loss}

Conventional reconstruction penalties (such as $\ell_1$ or standard perceptual metrics) operate under the assumption that the reference and generated images are perfectly aligned in physical space. This prerequisite completely breaks down in the context of handwriting synthesis. In this specific domain, stylistic characteristics (including stroke morphology, overall inclination, and curve geometry) manifest independently of their exact spatial coordinates. Furthermore, discrepancies in the underlying textual content introduce additional alignment complexities that standard losses cannot resolve. To overcome this structural limitation, we incorporate the Frequency Distribution Loss (FDL) \citep{Ni_2024FDL}. Following the methodological approach established by the FW-GAN architecture \citep{TONGDANGKHOA2026130175}, this metric is evaluated directly on \textit{high-level feature representations}. The fundamental concept involves interpreting the visual data as a collection of semantic elements, evaluating their overall statistical distributions while completely disregarding strict spatial coordinates. We derive these robust feature sets using a predetermined network operator $\Phi$, which specifically corresponds to the $l$-th layer of the shared feature extraction backbone previously detailed in Section~\ref{subsec:style_writer_modules}. This relationship is defined mathematically as:
\begin{equation}
	A = \{ a_1, a_2, \dots, a_N \} = \Phi^l(x), \quad
	B = \{ b_1, b_2, \dots, b_N \} = \Phi^l(G(y, E(x))),
\end{equation}
In this context, $x$ denotes the authentic reference picture, while $G(y, E(x))$ represents the corresponding reconstruction synthesized using the target content $y$ and the extracted style $E(x)$. Subsequently, we execute a Discrete Fourier Transform (DFT) across the spatial dimensions to separate every feature map into its respective amplitude ($\mathcal{A}$) and phase ($\mathcal{P}$) variables. These transformed frequency distributions are then evaluated against each other utilizing the Sliced Wasserstein Distance (SWD) metric:
\begin{equation}
	\mathcal{L}_{\text{FDL}} = \mathrm{SW}(\mathcal{A}(A), \mathcal{A}(B)) + \lambda \cdot \mathrm{SW}(\mathcal{P}(A), \mathcal{P}(B)).
\end{equation}
Within this mathematical framework, $\mathrm{SW}(\cdot, \cdot)$ stands for the Sliced Wasserstein Distance computed between two observed empirical distributions, whereas the coefficient $\lambda$ serves to calibrate the relative importance of the amplitude and phase variables.

\begin{figure}[!t]
	\centering
	\includegraphics[width=0.8\textwidth,height=5cm]{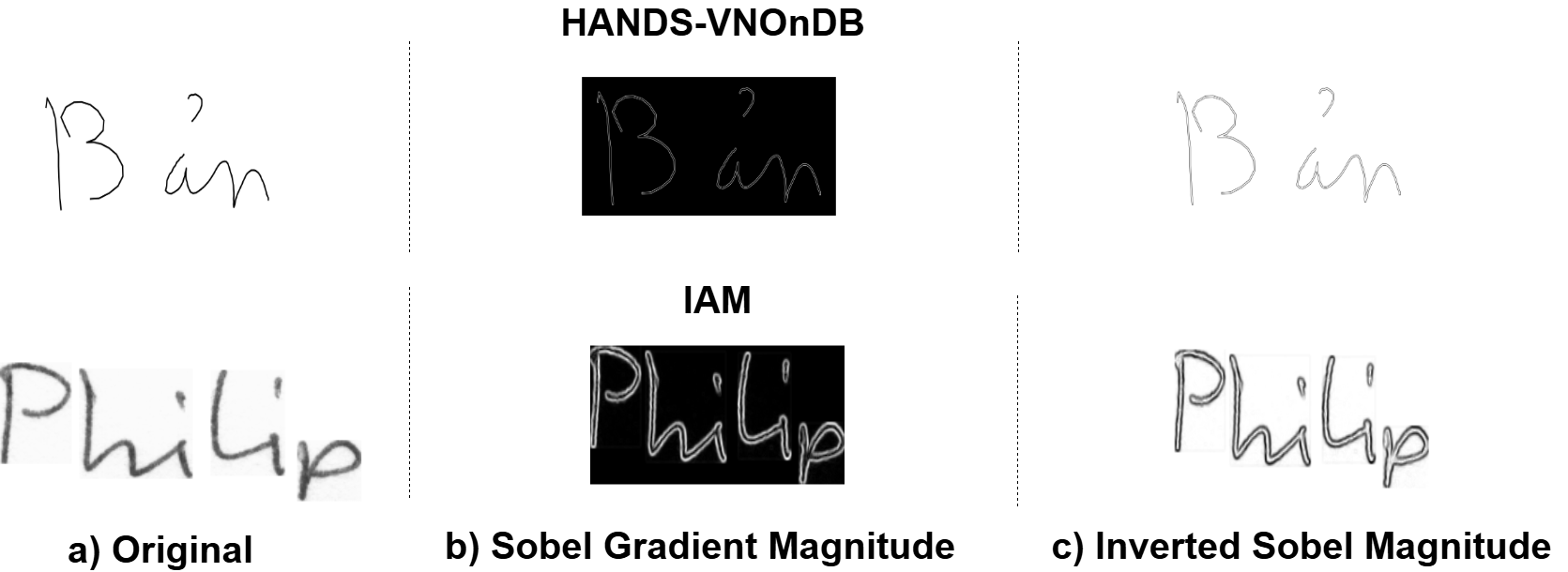}
	\caption{Visualization of Sobel edge magnitude maps.}
	\label{fig:sobel_edge}
\end{figure}

\subsubsection{Sobel-Regularized Edge Reconstruction Loss}

While FDL effectively handles spatial misalignment by comparing global spectral distributions, it does not focus on local high-frequency details such as stroke boundaries. This lack of explicit edge supervision can result in blurred character edges, as no direct constraint preserves sharp transitions at contours. In document enhancement, DocDiff~\citep{yang2023docdiff} addresses this limitation by employing a Laplacian kernel as a high-pass filter to extract high-frequency residuals along text edges, leveraging the observation that high-frequency information is primarily concentrated at edges. we introduce a Sobel-Regularized Edge Reconstruction Loss. Unlike the isotropic filter used in paper, our directional Sobel operator better suits offline handwriting images and improves performance, as verified by our experiments. We further use \(L_1\) over \(L_2\) to promote sparse gradients and sharper boundaries.

Given an image \(I\), we define its edge magnitude map using the Sobel operator:
\begin{equation}
	M_{\nabla}(I) = \sqrt{ (G_x \ast \bar{I})^2 + (G_y \ast \bar{I})^2 },
\end{equation}
where \(\bar{I}\) is the grayscale version, \(\ast\) denotes 2D spatial convolution, and \(G_x, G_y\) are the horizontal and vertical Sobel kernels. This operator extracts gradient information along both spatial axes, highlighting regions of sharp intensity changes that correspond to stroke boundaries (Figure~\ref{fig:sobel_edge} visualizes the resulting edge maps). Unlike DocDiff's residual-level supervision, which is designed to limit computational overhead in their diffusion model, our Sobel loss operates directly on edge maps, offering more explicit and interpretable guidance for stroke boundary preservation. We apply this loss to the reconstruction \(\hat{x} = G(z,y)\) against the ground-truth \(x\):
\begin{equation}
	\mathcal{L}_{edge} = \mathbb{E}_{x, y, z} \left[ \left\| M_{\nabla}(x) - M_{\nabla}(\hat{x}) \right\|_1 \right].
\end{equation}
where where \(\|\cdot\|_1\) is the \(L_1\) norm and the expectation \(\mathbb{E}_{x, y, z}\) is taken over the joint distribution of real images, content labels, and style vectors. Since Sobel is a fixed linear transform with no trainable parameters, \(\mathcal{L}_{edge}\) is fully differentiable, backpropagates directly to the generator. We integrate the new loss with other objectives (Sec.~\ref{sec:overall_objectives}). Our loss directly supervises edge sharpness, preserving the fine structural details essential for handwriting legibility and style consistency. Additionally, FDL handles global spectral matching, while ours acts as a directional spatial regularizer that enforces stroke fidelity, complementing distribution-based supervision with explicit geometric constraints.

\subsubsection{Spiral-Modulated Adversarial Loss}

While the standard adversarial loss encourages global visual realism, handwriting synthesis also depends on fine-grained structural attributes such as stroke continuity, character integrity, and curved stroke trajectories. To explicitly enforce realism in these structural details, we leverage the Spiral-Modulated Discriminator $D_{SP}$ described in~\ref{subsubsec:discriminator}, which processes input features through three parallel pathways: spatial ($X^{\text{spatial}}$), spiral ($X^{\text{spiral}}$), and spectral ($X^{\text{spectral}}$). The discriminator adaptively fuses these representations via independent gating mechanisms, enabling it to capture both local stroke patterns and global structural coherence. Given an input image $x$, the fused representation is computed as the gated combination of these three feature streams, allowing $D_{SP}$ to focus on the most informative structural cues for each input.

The Spiral-Modulated adversarial loss adopts the same hinge loss formulation as the standard adversarial loss:
\begin{equation} 
	\mathcal{L}_{SP} = \mathbb{E}_{x \sim \mathbf{X}} \left[ \max(0, 1 - D_{SP}(x)) \right] + \mathbb{E}_{\tilde{y} \sim \mathcal{C},\, z} \left[ \max(0, 1 + D_{SP}(G(\tilde{y}, z))) \right].
\end{equation}
where $D_{SP}(x)$ denotes the realism score produced by the spiral-modulated discriminator for a real image $x$, and $D_{SP}(G(\tilde{y}, z))$ is the score for a generated image $G(\tilde{y}, z)$.

For generator training, $\mathcal{L}_{SP}$ encourages the generator to produce images with structurally coherent strokes and sharp character edges. In our dual-discriminator framework, $\mathcal{L}_{adv}$ provides global structural supervision through the standard spatial discriminator, while $\mathcal{L}_{SP}$ complements it with targeted supervision on structural integrity, ensuring handwriting outputs are both visually realistic and structurally intact.

\subsubsection{Overall Objectives}
\label{sec:overall_objectives}
We optimize our architecture utilizing a min-max adversarial paradigm, which harmonizes the competing goals of the generative and discriminative networks while incorporating supplementary supervisory tasks. The comprehensive optimization sequence is detailed in the subsequent paragraphs.

\paragraph{\textbf{Discriminators and Auxiliary Modules.}} Throughout the maximization stage, we individually refine the parameters of the conventional discriminator $D$, the specialized MLP-based Discriminator $D_{SP}$, the recognition module $R$, and the author classification network $W$. The specific optimization targets for these independent components are formulated as follows:
\begin{equation}
	\mathcal{L}_D = -\mathcal{L}_{adv}, \quad
	\mathcal{L}_{D_{SP}} = -\mathcal{L}_{SP}, \quad
	\mathcal{L}_R = \mathcal{L}_{R}^D, \quad
	\mathcal{L}_W = \mathcal{L}_{W}^D,
\end{equation}

\paragraph{\textbf{Generator and Style Encoder.}} Conversely, during the minimization phase, the generative component $G$ and the encoding module $E$ undergo simultaneous parameter updates aimed at reducing the following aggregated penalty:
\begin{equation}
	\mathcal{L}_{G,E} = \mathcal{L}_{adv} + \mathcal{L}_{SP} + \lambda_{R} \mathcal{L}_{R}^G + \lambda_{W} \mathcal{L}_{W}^G + \lambda_{style} \mathcal{L}_{style} + \lambda_{kl} \mathcal{L}_{kl} + \lambda_{FDL}\mathcal{L}_{FDL} + \lambda_{edge}\mathcal{L}_{edge}.
\end{equation}
Within this equation, every $\lambda$ variable acts as a scaling coefficient designed to calibrate the relative influence of its specific error metric. Working in unison, these individual components regulate distinct dimensions of the artificial penmanship process, encompassing aesthetic authenticity, character readability, strict stylistic adherence, and the structural normalization of the latent domain.

\section{Experiments}\label{sec:experiments}

We thoroughly investigate the capabilities of our newly introduced SpiS-GAN architecture by conducting an extensive series of experimental trials aimed at measuring both the visual fidelity of the synthesized outputs and their practical utility in applied scenarios. This segment initially outlines our testing methodology alongside specific technical deployment parameters. Subsequently, it presents a detailed performance comparison against leading state-of-the-art methodologies, systematic ablation analyses, and rigorous cross-language adaptability evaluations. Furthermore, we explore the positive impact of our artificial data on subsequent downstream handwriting recognition systems and critically assess the computational efficiency of the framework upon deployment.

\subsection{Design of Evaluation Procedure} \label{sub:evaluation_procedure}

Evaluating progress in generating handwritten text has historically been difficult due to the absence of a standardized testing framework. Because previous studies relied on widely differing evaluation techniques, it is nearly impossible to make direct comparisons or draw concrete conclusions across the literature. To resolve this inconsistency, our research rigorously implements the structured, reproducible evaluation methodology introduced by the FW-GAN model \cite{TONGDANGKHOA2026130175}, which was explicitly engineered to address the unique requirements of handwriting synthesis.

Although this testing framework is adaptable and functions effectively across various data collections, we explain its mechanics using the IAM dataset, given its overwhelming popularity in related academic research. This specific dataset comprises roughly 62,857 images of handwritten English words, collected from 500 distinct individuals. We disregard the original dataset partition (which divides authors into training, validation, and testing segments). Instead, we adopt the optimized split that has become standard practice in contemporary studies like HWT \cite{bhunia2021handwriting} and VATr \cite{pippi2023handwritten}. This configuration allocates 339 authors specifically for system training, holding back the remaining 161 individuals exclusively for testing purposes. To rigorously assess the generative capabilities of our model, we establish five highly controlled evaluation scenarios that test its ability to generalize both linguistically and stylistically:

\begin{itemize}
	\item \textbf{Test Set Replication:} Within this specific trial, the system rebuilds the entire test set by generating every single target word precisely as it exists in the actual test data. For every artificial image produced, the model extracts the required style from reference pictures written by the exact same author containing that specific word. This allows us to strictly measure the network's capacity to flawlessly recreate both the required content and the unique stylistic quirks of the original writer.
	\item \textbf{IV-S (In-Vocabulary, Seen Style):} The model generates words it encountered during training, applying handwriting styles from authors it also studied during training.
	\item \textbf{IV-U (In-Vocabulary, Unseen Style):} The model generates familiar words, but must render them using the handwriting styles of entirely unfamiliar authors.
	\item \textbf{OOV-S (Out-of-Vocabulary, Seen Style):} The model must render completely new words that it never saw during training, utilizing the familiar handwriting styles of known authors.
	\item \textbf{OOV-U (Out-of-Vocabulary, Unseen Style):} The model faces the ultimate challenge: rendering entirely new words using the handwriting styles of completely unfamiliar authors, forcing the network to generalize across both dimensions simultaneously.
\end{itemize}

To construct these final four testing categories, we compile two distinct word lists: one comprising vocabulary items found within the training data, and a second containing entirely new words pulled from an independent English text corpus. During each specific testing scenario, the model synthesizes approximately 25,000 artificial word pictures. We subsequently compare these generated outputs against authentic handwriting samples utilizing strict numerical metrics.

To ensure an unbiased measurement of output quality, we deploy a combination of perceptual and style-focused evaluation metrics. Primarily, we document the \textit{Fréchet Inception Distance} (FID) and the \textit{Kernel Inception Distance} (KID). These two metrics are the standard tools utilized throughout generative research to quantify the mathematical similarity between the distributions of authentic and artificial image sets \cite{kang2020ganwriting, gan2021higan, gan2022higan+, bhunia2019handwriting, pippi2023handwritten, nikolaidou2024diffusionpen, one-dm2024,TONGDANGKHOA2026130175}. Both of these techniques rely on high-level feature sets extracted from a pre-trained InceptionV3 architecture. Specifically, FID computes the Fréchet distance comparing two multivariate Gaussian models applied to the feature data, whereas KID calculates the squared maximum mean discrepancy utilizing a polynomial kernel mapping. It is well known that these metrics possess inherent flaws when evaluating handwriting, primarily because the underlying Inception network was trained on standard photographs (ImageNet), creating a significant domain mismatch when analyzing text. Despite this flaw, we report FID and KID to maintain consistency with historical literature. To compensate for these perceptual limitations and directly counteract the domain shift problem, we integrate the recently developed \textit{Handwriting Distance} (HWD) metric \cite{pippi2023hwd}. In contrast to FID and KID, the HWD metric utilizes a VGG16 network explicitly trained on a massive collection of handwritten text, calculating direct Euclidean distances between the authentic and generated data. This yields a measurement that is significantly more sensitive to stylistic nuances and perfectly aligned with the handwriting domain, serving as a critical counterpart to the traditional metrics. For maximum clarity, we evaluate all these metrics using two distinct approaches: first, by calculating them per individual writer and averaging the final results, and second, by evaluating them globally across the complete test collection.

\subsection{Implementation Details}\label{subsec:setting}

We construct the SpiS-GAN architecture utilizing the PyTorch library\footnote{\url{https://pytorch.org/}} and execute all computational experiments on a single NVIDIA A100 graphical processing unit. For network optimization, we employ the Adam algorithm \cite{kingma2014adam}, establishing an initial learning rate of $0.0002$ alongside momentum parameters configured to $(\beta_1, \beta_2) = (0.5, 0.999)$. Once the training process reaches the 35\textsuperscript{th} epoch, we implement a linear decay schedule to systematically reduce the learning rate over time.

Regarding the mathematical loss weighting, we assign static numerical values to three particular components: $\lambda_{\text{KL}} = 0.0001$ dedicated to the KL divergence, $\lambda_{\text{FDL}} = 1$ managing the frequency distribution penalty, and $\lambda_{\text{edge}} = 1$ controlling the sobel-edge reconstruction loss. We intentionally decline to assign fixed values for all remaining auxiliary loss coefficients. Instead, our architecture automatically calibrates these specific weights dynamically throughout the training phase utilizing a specialized gradient balancing algorithm. This strategic approach guarantees that no individual optimization target ever disproportionately dominates the learning process.

	\subsection{Benchmarking Against State-of-the-Art Methods}

To evaluate the efficacy of our introduced architecture, we perform comprehensive evaluations against multiple prominent methodologies specialized in style-conditioned handwritten text generation. The selected baseline architectures encompass the foundational GANWriting \cite{kang2020ganwriting}, the transformer-based framework HWT \cite{bhunia2021handwriting}, VATr \cite{pippi2023handwritten}, and the subsequent developments HiGAN \cite{gan2021higan} alongside HiGAN+ \cite{gan2022higan+}. Furthermore, we benchmark against cutting-edge diffusion-based networks, notably DiffusionPen \cite{nikolaidou2024diffusionpen} and One-DM \cite{one-dm2024}, in addition to the preceding GAN-based pinnacle FWGAN \citep{TONGDANGKHOA2026130175}. This comprehensive selection allows us to contextualize our contributions across both the adversarial and diffusion generative landscapes.

Aligning with historical literature, contemporary networks utilize varying structural setups that yield final images with vertical resolutions of either 32 or 64 pixels. To facilitate an equitable analysis, we test our system across both dimensional standards. The 32-pixel iteration strictly adheres to the structural blueprint detailed previously. Conversely, the 64-pixel adaptation integrates a supplementary SpiralGBlock module at the terminal end of the generator to effectively expand the spatial resolution from 32 to 64 pixels. As a result, every generated picture is strictly normalized to match one of these specific vertical dimensions. Within the 32-pixel configuration, individual characters receive an allocation of 16 pixels. For the 64-pixel format, this character allocation increases to 32 pixels. To properly handle textual sequences of varying lengths, we systematically pad any images falling below the 128-pixel width threshold. Meanwhile, pictures exceeding 128 pixels in the smaller format, or 256 pixels in the larger format, undergo a resizing operation to achieve exact target dimensions of $32 \times 128$ or $64 \times 256$ respectively.

To guarantee a strict and impartial testing environment, we faithfully execute the methodological protocol formulated by FWGAN. Regarding baseline frameworks that provide accessible pretrained parameters optimized on matching data partitions (such as HWT and VATr), we directly utilize their official distributions. For architectures lacking accessible pretrained models (including HiGAN and HiGAN+), we fully retrain them from the ground up utilizing identical dataset divisions to ensure absolute parity. This careful strategy guarantees highly consistent testing conditions, effectively eliminating any variations stemming from distinct data preparation or optimization routines. For a specific group of baselines, encompassing GANWriting evaluated on English and Vietnamese datasets, alongside DiffusionPen tested on VietNamese, we directly cite the performance metrics documented by FWGAN \citep{TONGDANGKHOA2026130175}. We present rigorous numerical comparisons utilizing three recognized indicators: Fréchet Inception Distance (FID), Kernel Inception Distance (KID), and Handwriting Distance (HWD). Throughout every configured testing scenario, our introduced SpiS-GAN architecture consistently surpasses all existing methods concerning both the FID and KID scores, clearly demonstrating enhanced visual authenticity and strict stylistic coherence.

\subsection{Quantitative Evaluation}\label{sec:quantitative}

Table~\ref{tab:FID-KID} offers a detailed side-by-side comparison of handwriting synthesis quality across multiple state-of-the-art methods, using FID and KID metrics at two different resolutions. Our model establishes a new state-of-the-art in both 32-pixel and 64-pixel height settings, achieving the lowest scores on both evaluation metrics in each category.

At the 32-pixel resolution, our model attains top performance with a FID of 4.37 and KID of 0.06, demonstrating the strongest resemblance to genuine handwriting among all evaluated approaches. By contrast, HiGAN records a considerably higher FID of 15.017, highlighting a substantial performance gap. Transformer-based models such as HWT (FID: 14.115, KID: 0.51) and VATr (FID: 13.277, KID: 0.45) outperform HiGAN but still fall noticeably behind our results. Impressively, our FID is less than half that of FWGAN (6.73), the second-best model, and our KID is similarly reduced by over 50\%.

For the 64-pixel setting, our model preserves this advantage with a FID of 4.58 and KID of 0.08, comfortably surpassing all competing methods at this higher resolution. Among the baselines, One-DM~\cite{one-dm2024} achieves a FID of 15.59 and KID of 0.75, while HiGAN+~\cite{gan2022higan+} attains a FID of 10.17 and KID of 0.45, making it the strongest competitor in this category. Yet even HiGAN+, the top-performing baseline at 64 pixels, records a FID more than double ours. Across both resolutions, our method consistently produces FID scores below half and KID scores substantially lower than those of the next best models, confirming its reliable and superior ability to generate high-quality handwriting regardless of image resolution.

Table~\ref{tab:oov} presents FID scores across four test scenarios specifically designed to evaluate both vocabulary generalization and style robustness. Our model achieves the lowest FID scores in all four conditions at both resolutions, reflecting strong and consistent performance.

For 32-pixel images, in the IV-S setting, our method surpasses FWGAN (25.01), the nearest competitor, by 2.12 points, and outperforms HiGAN (35.02) by more than 12 points. In the IV-U scenario, which assesses style generalization to unseen writers using familiar vocabulary, our model scores 26.33, exceeding FWGAN (27.94) by 1.61 points and VATr (28.14) by 1.81 points, while also achieving over 12-point improvements over HiGAN (39.12). In the OOV-S setting, where novel words are generated in familiar styles, our approach leads FWGAN by 2.14 points and VATr by 3.61 points. Most notably, in the most challenging OOV-U setting—where both words and styles are entirely unseen—our model achieves 26.76, surpassing FWGAN (27.17) by 0.41 points, VATr (29.51) by 2.75 points, HWT (29.68) by 2.92 points, and HiGAN (40.35) by an impressive 13.59 points. Although FWGAN remains competitive in certain OOV scenarios, no single baseline—whether GAN-based or transformer-based—maintains consistent superiority across all settings. In contrast, our method maintains stable and low FID scores across every condition, demonstrating its resilience to domain shifts and its ability to generate high-fidelity handwriting from familiar contexts to the most challenging OOV-U case.

For 64-pixel images, our model continues to produce strong results across all scenarios. In the IV-S setting, we achieve 31.27, outperforming HiGAN+ (32.15), One-DM (35.18), and earlier approaches like DiffusionPen (42.26) and GanWriting (45.49). In IV-U, our model attains 28.52, the top score, narrowly surpassing HiGAN+ (33.07) and One-DM (32.95), with substantial margins over DiffusionPen (42.16) and GanWriting (48.58). For OOV-S, we record 32.22, leading all competitors including HiGAN+ (37.48), One-DM (35.55), DiffusionPen (39.07), and GanWriting (42.64). In the challenging OOV-U case, our method scores 29.40, surpassing HiGAN+ (31.55) and One-DM (32.92), while also comfortably outperforming earlier generation methods. At 64-pixel resolution, our model consistently maintains low FID scores, confirming its stable generalization capabilities.

These findings underscore the strength of our proposed model in tackling real-world handwriting generation challenges, where both the textual content and the writer's style can vary significantly. The capacity to generalize under such diverse conditions is essential for practical deployment, and our results demonstrate that our model provides a robust and adaptable solution for style-driven handwritten text generation.

\begin{table}[!t]
	\centering
	\caption{Quantitative comparison of handwriting synthesis quality (FID and KID) on the IAM test set. Lower is better. Results marked with * are taken directly from FWGAN~\cite{TONGDANGKHOA2026130175}.}
	\label{tab:FID-KID}
	\begin{tabular}{lcc}
		\toprule
		\textbf{Method} & \textbf{FID} & \textbf{KID} \\
		\midrule
		\multicolumn{3}{l}{\textbf{32-pixel height}} \\
		\midrule
		HiGAN~\cite{gan2021higan}      & 15.017 & 0.91 \\
		HWT~\cite{bhunia2021handwriting} & 14.115 & 0.51 \\
		VATr~\cite{pippi2023handwritten} & 13.277 & 0.45 \\
		FWGAN~\cite{TONGDANGKHOA2026130175} & 6.730 & 0.22 \\
		\textbf{Ours-32}                  & \textbf{4.37} & \textbf{0.06} \\
		\midrule
		\multicolumn{3}{l}{\textbf{64-pixel height}} \\
		\midrule
		GanWriting*~\cite{kang2020ganwriting}  & 31.20 & 1.50  \\
		DiffusionPen~\cite{nikolaidou2024diffusionpen}        & 28.92 & 1.58 \\
		One-DM~\cite{one-dm2024}        & 15.59 & 0.75 \\
		HiGAN+~\cite{gan2022higan+}    & 10.17 & 0.45  \\
		\textbf{Ours-64}               & \textbf{4.58} & \textbf{0.08} \\
		\bottomrule
	\end{tabular}
\end{table}

\begin{table}[!t]
	\centering
	\caption{FID scores isolating vocabulary (In/Out) and style (Seen/Unseen) generalization on IAM (25,000 samples per setting). Results marked with * are taken directly from FWGAN~\cite{TONGDANGKHOA2026130175}.}
	\label{tab:oov}
	\begin{tabular}{lcccc}
		\toprule
		\textbf{Method} & \textbf{IV-S} & \textbf{IV-U} & \textbf{OOV-S} & \textbf{OOV-U} \\
		\midrule
		\multicolumn{5}{l}{\textbf{32-pixel height}} \\
		\midrule
		HiGAN~\cite{gan2021higan}             & 35.02 & 39.12 & 33.14 & 40.35  \\
		HWT~\cite{bhunia2021handwriting}      & 26.94 & 29.16 & 26.47 & 29.68 \\
		VATr~\cite{pippi2023handwritten}      & 26.23 & 28.14 & 27.02 & 29.51 \\
		FWGAN~\cite{TONGDANGKHOA2026130175}   & 25.01 & 27.94 & 25.55 & 27.17 \\
		\textbf{Ours-32}                     & \textbf{22.89} & \textbf{26.33} & \textbf{23.41} & \textbf{26.76}\\
		\midrule
		\multicolumn{5}{l}{\textbf{64-pixel height}} \\
		\midrule
		GanWriting*~\cite{kang2020ganwriting}  & 45.49 & 48.58 & 42.64 & 43.92  \\
		DiffusionPen~\cite{nikolaidou2024diffusionpen} & 42.26 & 42.16 & 39.07 & 39.47 \\
		HiGAN+~\cite{gan2022higan+}           & 32.15 & 33.07 & 37.48 & 31.55 \\
		One-DM~\cite{one-dm2024}              & 35.18 & 32.95 & 35.55 & 32.92 \\
		\textbf{Ours-64}                     & \textbf{31.27} & \textbf{28.52} & \textbf{32.22} & \textbf{29.40} \\
		\bottomrule
	\end{tabular}
\end{table}

\subsection{Enhancing HTR Performance via Synthetic Data}

To rigorously evaluate the practical utility of our proposed generative framework, we analyze its influence on the accuracy of downstream Handwritten Text Recognition (HTR) systems, particularly within environments constrained by limited authentic data. Ultimately, the primary objective of artificial penmanship generation extends beyond merely producing aesthetically convincing pictures; it must also actively support subsequent recognition algorithms by supplying highly informative supplemental training datasets.

To accurately simulate a low-resource scenario, we initially establish a baseline recognition model optimized on a restricted collection of merely 5,000 previously unseen authentic handwriting specimens extracted from the IAM database. Throughout this specific evaluation, we implement a Transformer-based optical character recognition architecture modeled after the established TrOCR methodology \cite{Li_2023TrOCR}. Subsequently, we drastically expand this foundational dataset by incorporating 25,000 artificial images generated by every competing framework alongside our own. These generated pictures utilize a randomized lexicon directly derived from the initial authentic sample pool. This newly expanded corpus (comprising exactly 5,000 real and 25,000 artificial images) is then deployed to completely retrain the recognition network. By maintaining a strictly constant set of authentic images and exclusively altering the origin of the synthetic augmentation, we can objectively isolate and quantify the specific performance improvements provided by each individual generation technique. We meticulously measure the final recognition accuracy utilizing three universally recognized evaluation metrics:

\begin{itemize}
	\item \textbf{Character Error Rate (CER):} This metric calculates the specific proportion of character-level inaccuracies, encompassing erroneous substitutions, unwanted insertions, and missing deletions, directly compared against the total character count present within the ground-truth annotations.
	\item \textbf{Word Error Rate (WER):} This value tracks the total percentage of improperly transcribed words, providing a macro-level perspective regarding the overall recognition quality.
	\item \textbf{Normalized Edit Distance (NED):} This calculation evaluates the standardized mathematical distance separating the predicted output from the target sequence, wherein a reduced numerical value signifies a significantly higher degree of transcription accuracy.
\end{itemize}

As shown in Table~\ref{HTR}, every synthetic augmentation strategy leads to noticeable performance gains over the baseline trained exclusively on 5,000 real images, confirming that synthetic handwriting effectively makes up for limited real data.

For 32-pixel height images, our method achieves the lowest error rates among all competing approaches across all three metrics, with a CER of 10.03, NED of 9.99, and WER of 27.71. Compared to the strongest baseline competitor FWGAN (CER 10.18, NED 10.19, WER 28.38), our method delivers reductions of 0.15 CER points, 0.20 NED points, and 0.67 WER points. VATr attains CER 10.52, NED 10.32, and WER 29.15, while HWT records CER 10.75, NED 10.70, and WER 29.31. HiGAN~\cite{gan2021higan} logs CER 10.98, NED 10.51, and WER 30.02. These gains are consistent across both character- and word-level metrics, suggesting that our generated samples boost fine-grained character recognition while also strengthening word-level prediction.

For 64-pixel height images, our method shows even more substantial improvements, achieving CER 9.01, NED 8.83, and WER 26.02—the lowest error rates in this resolution category. The nearest competitor, HiGAN+, reaches CER 9.45, NED 9.58, and WER 27.33, which corresponds to improvements of 0.44 CER points, 0.75 NED points, and 1.31 WER points for our method. Other methods demonstrate varying degrees of effectiveness. One-DM~\cite{one-dm2024} records CER 10.76, NED 10.65, and WER 30.26, while DiffusionPen shows CER 11.86, NED 11.52, and WER 31.81. Notably, GanWriting~\cite{kang2020ganwriting} yields the weakest results with CER 17.62, NED 16.73, and WER 44.38, indicating that its synthetic samples may introduce noise instead of helpful training information. The significant WER reductions achieved by our method at both resolutions suggest that our model generates more semantically coherent handwriting, enabling the HTR model to better capture complete word structures rather than just individual characters.

This consistent performance across multiple metrics and both resolutions reinforces that our synthesis approach produces data that is both visually convincing and practically valuable for training robust recognition systems. The superior HTR performance, combined with the lower FID and KID scores reported earlier, provides strong evidence that our method generates high-quality synthetic handwriting well-suited for data augmentation and low-resource applications.

\begin{table}[!t]
	\centering
	\caption{HTR performance on IAM using 5,000 real images augmented with 25,000 synthetic images. Lower is better. Results marked with * are taken directly from FWGAN~\cite{TONGDANGKHOA2026130175}.}
	\label{HTR}
	\begin{tabular}{lccc}
		\toprule
		\textbf{Method} & \textbf{CER} & \textbf{NED} & \textbf{WER} \\
		\midrule
		\multicolumn{4}{l}{\textbf{32-pixel height}} \\
		\midrule
		5000 real images     & 12.25 & 11.95 & 32.81 \\
		HiGAN~\cite{gan2021higan}                & 10.98 & 10.51 & 30.02 \\
		HWT~\cite{bhunia2021handwriting}         & 10.75 & 10.70 & 29.31 \\
		VATr~\cite{pippi2023handwritten}         & 10.52 & 10.32 & 29.15 \\
		FWGAN~\cite{TONGDANGKHOA2026130175}          & 10.18 & 10.19 & 28.38 \\
		\textbf{Ours-32}     & \textbf{10.03} & \textbf{9.99} & \textbf{27.71} \\
		\midrule
		\multicolumn{4}{l}{\textbf{64-pixel height}} \\
		\midrule
		5000 real images     & 10.81 & 10.66 & 30.31 \\
		GanWriting*~\cite{kang2020ganwriting}  & 17.62 & 16.73 & 44.38 \\
		DiffusionPen~\cite{nikolaidou2024diffusionpen}  & 11.86 & 11.52 & 31.81 \\
		One-DM~\cite{one-dm2024}                & 10.76 & 10.65 & 30.26 \\
		HiGAN+~\cite{gan2022higan+}           & 9.45 & 9.58 & 27.33 \\
		\textbf{Ours-64}     & \textbf{9.01} & \textbf{8.83} & \textbf{26.02} \\
		\bottomrule
	\end{tabular}
\end{table}

\subsection{Ablation Study}

To evaluate the individual contributions of each component in our SpiS-GAN framework, we conduct a series of controlled experiments on the IAM dataset. We begin with a baseline configuration (A) that employs a BigGAN generator along with the Frequency Distribution Loss (FDL). We then progressively incorporate the Star-Spiral Blocks into the generator, followed by the Spiral-Modulated Discriminator, and finally the Sobel-Regularized Edge Loss. We assess image quality using Fréchet Inception Distance (FID) and measure recognition performance using Character Error Rate (CER), Normalized Edit Distance (NED), and Word Error Rate (WER).

Table~\ref{tab:ablation-study} reports the quantitative outcomes. The baseline (A) achieves an FID of 11.37, a CER of 11.20, an NED of 10.99, and a WER of 30.08. Adding the Star-Spiral Blocks (B) reduces the FID to 5.91 and improves the CER to 10.98. This indicates a substantial boost in overall image quality and basic text structure. Incorporating the Spiral-Modulated Discriminator (C) further lowers the FID to 4.61 and the CER to 10.26. This confirms that evaluating full-resolution pen strokes helps prevent structural degradation in character shapes. Finally, the addition of the Sobel-Regularized Edge Loss (D) yields our complete model (Ours-32), which achieves the best performance across all metrics: an FID of 4.37, a CER of 10.18, an NED of 9.99, and a WER of 27.71.

\begin{figure}[!t]
	\centering
	\includegraphics[width=0.70\linewidth]{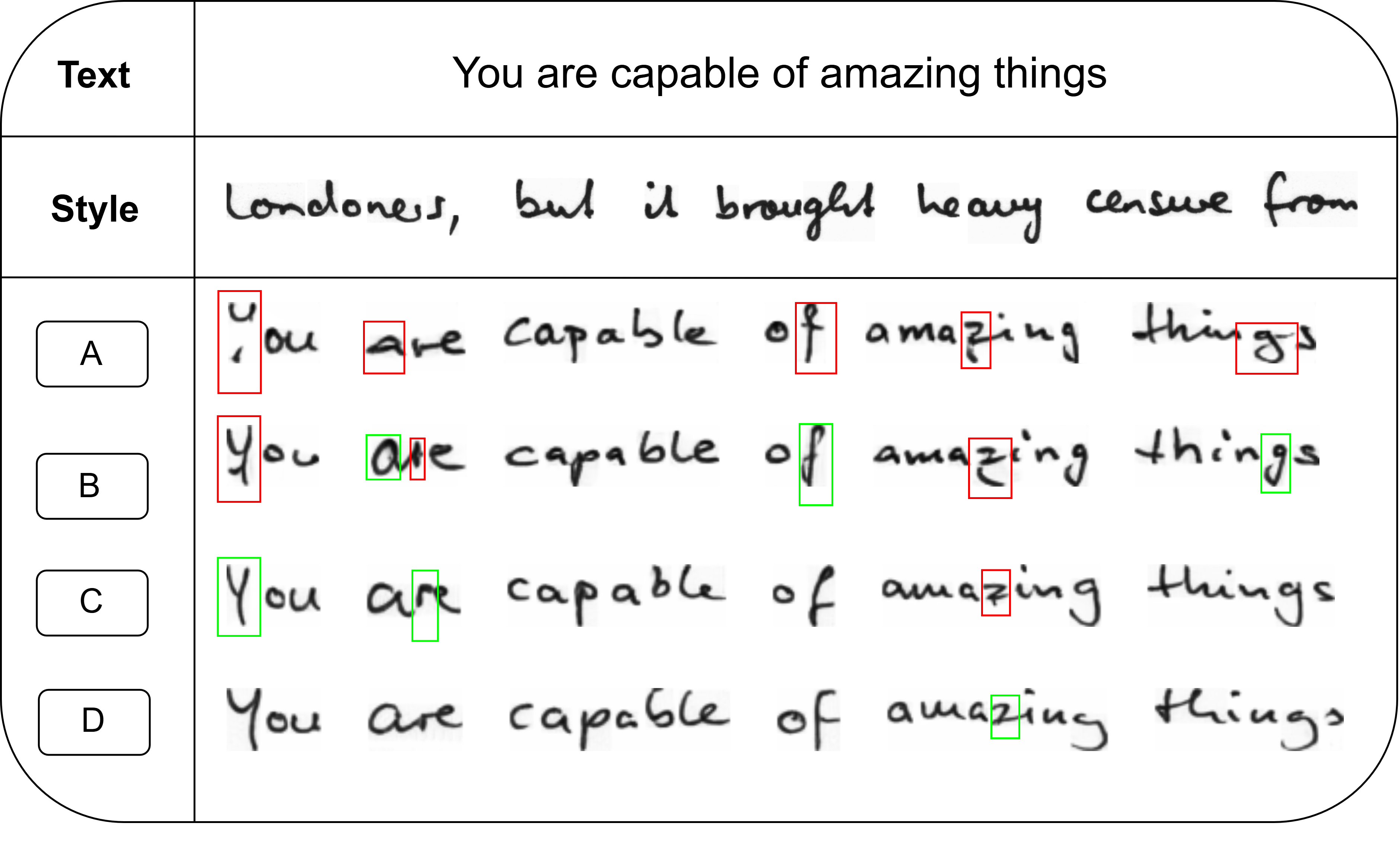}
	\caption{Visual progression of the ablation study. Red boxes indicate faint/blurred edges; green boxes highlight successful structural and stylistic correction. Models A through D align with Table~\ref{tab:ablation-study}.}
	\label{fig:ablation_study_visualization}
\end{figure}

Visual inspection of the samples in Figure~\ref{fig:ablation_study_visualization} supports these quantitative findings. The baseline (A) struggles to produce solid ink, resulting in faint, broken, and poorly formed strokes (highlighted by red boxes on 'Y', 'a', 'f', 'z', and 'g'). With the Star-Spiral Blocks (B), the text becomes noticeably sharper, and basic shape errors are corrected (green boxes on 'a', 'f', 'g'). However, the overall writing style still lacks the proper ink thickness and natural cursive flow (red boxes remain on 'Y', 'r', and 'z'). Introducing the Spiral-Modulated Discriminator (C) resolves these stylistic and contrast issues, successfully correcting ink weight and stroke flow for characters like 'Y' and 'r' (green boxes). Nonetheless, minor shape imperfections persist on complex connected letters (red box on 'z'). Finally, applying the SELoss (D) produces crisp stroke boundaries, completely eliminating residual blur and errors on intricate characters such as the 'z' (green box), achieving a near-perfect match with the target style.

\begin{table}[!t]
	\centering
	\caption{Ablation study on IAM. We track generative fidelity (FID) and recognition performance (CER, NED, WER).}
	\label{tab:ablation-study}
	\makebox[\textwidth][c]{%
		\begin{tabular}{lcccc}
			\toprule
			\textbf{Model} & \textbf{FID} & \textbf{CER} & \textbf{NED} & \textbf{WER} \\
			\midrule
			Baseline (BigGAN) + FDL (A) & 11.37 & 11.20 & 10.99 & 30.08 \\
			(A) + Star-Spiral Blocks (Gen) (B) & 5.91 & 10.98 & 10.51 & 28.54 \\
			(B) + Spiral-Modulated Discriminator (C) & 4.61 & 10.26 & 10.11 & 28.23 \\
			(C) + Sobel-Regularized Edge Reconstruction Loss (D) & \textbf{4.37} & \textbf{10.18} & \textbf{9.99} & \textbf{27.71} \\
			\bottomrule
		\end{tabular}%
	}
\end{table}

\subsection{Cross-Language Generalization: Vietnamese Handwriting}

To assess the ability of SpiS-GAN to generalize across different writing systems and its robustness in handling diverse scripts, we carry out additional experiments on Vietnamese following the same experimental setup described in~\ref{subsec:setting}. For this evaluation, we use the HANDS-VNOnDB dataset~\cite{HANDS-VNOnDB}, which contains 7,296 handwritten lines with over 480,000 strokes and more than 380,000 characters. Following the official data split, we train the model on handwriting from 106 writers and test it on 34 previously unseen writers.

Table~\ref{tab:FID-KID-Vietnamese} shows that SpiS-GAN consistently achieves the best results among all evaluated models at both 32-pixel and 64-pixel resolutions. At 32-pixel height, our method records the lowest FID (3.67) and KID (0.14), significantly outperforming strong baselines such as FWGAN (FID: 5.72, KID: 0.29), HWT (FID: 9.15, KID: 0.69), and HiGAN (FID: 11.36, KID: 0.69). At 64-pixel height, our model also exceeds recent high-resolution approaches like HiGAN+ and DiffusionPen, setting a new state-of-the-art with a FID of 4.35 and KID of 0.20. Compared to the best baseline at 64 pixels, HiGAN+ (FID: 7.05, KID: 0.40), SpiS-GAN lowers FID by 2.70 points and cuts KID in half (from 0.40 to 0.20).

These consistent gains at both resolutions indicate that our model produces handwriting that appears more realistic and better matches the true data distribution. Vietnamese handwriting presents unique difficulties not found in English, such as the frequent use of diacritical marks, compound vowels, and tone markers, all of which add extra spatial and stylistic complexity. Despite these added challenges, SpiS-GAN successfully captures fine-grained structural and stylistic details, generating handwriting that is visually convincing and stylistically consistent across different writers. The fact that our model performs so well on a language that is structurally very different from English highlights the robustness of our approach and confirms its ability to generalize high-quality handwriting synthesis well beyond the language it was trained on.

\begin{table}[!t]
	\centering
	\caption{Performance on the HANDS-VNOnDB dataset (FID and KID). Lower is better. Results marked with * are taken directly from FWGAN~\cite{TONGDANGKHOA2026130175}.}
	\label{tab:FID-KID-Vietnamese}
	\begin{tabular}{lcc}
		\toprule
		\textbf{Method} & \textbf{FID} & \textbf{KID} \\
		\midrule
		\multicolumn{3}{l}{\textbf{32-pixel height}} \\
		\midrule
		VATr~\cite{pippi2023handwritten}       & 23.85 & 2.70 \\
		HiGAN~\cite{gan2021higan}              & 11.36 & 0.69 \\
		HWT~\cite{bhunia2021handwriting}       & 9.15 & 0.69 \\
		FWGAN~\cite{TONGDANGKHOA2026130175}        & 5.72 & 0.29 \\
		\textbf{Ours-32}                   & \textbf{3.67} & \textbf{0.14} \\
		\midrule
		\multicolumn{3}{l}{\textbf{64-pixel height}} \\
		\midrule
		GanWriting*~\cite{kang2020ganwriting}  & 31.20 & 1.50  \\
		DiffusionPen*~\cite{nikolaidou2024diffusionpen}    & 21.25 & 2.41 \\
		HiGAN+~\cite{gan2022higan+}    & 7.05 & 0.40 \\
		\textbf{Ours-64}             & \textbf{4.35} & \textbf{0.20} \\
		\bottomrule
	\end{tabular}
\end{table}

\subsection{Writer-Level Evaluation}
\label{subsec:writer_evaluation}

\begin{table}[!t]
	\centering
	\caption{Writer-level FID, KID, and HWD on the IAM dataset. Lower indicates better writer identity preservation. Results marked with * are taken directly from FWGAN~\cite{TONGDANGKHOA2026130175}.}
	\label{tab:FID-KID-IAM-writer}
	\begin{tabular}{lccc}
		\toprule
		\textbf{Method} & \textbf{FID} & \textbf{KID} & \textbf{HWD} \\
		\midrule
		\multicolumn{4}{l}{\textbf{32-pixel height}} \\
		\midrule
		HiGAN~\cite{gan2021higan}      & 15.02 & 0.90 & 0.78 \\
		HWT~\cite{bhunia2021handwriting} & 14.12 & 0.50 & 0.90 \\
		VATr~\cite{pippi2023handwritten} & 13.28 & 0.45 & 0.73 \\
		FWGAN~\cite{TONGDANGKHOA2026130175} & 6.73 & 0.22 & 0.57 \\
		\textbf{Ours-32}    & \textbf{4.37} & \textbf{0.06} & \textbf{0.51} \\
		\midrule
		\multicolumn{4}{l}{\textbf{64-pixel height}} \\
		\midrule
		GANwriting*~\cite{kang2020ganwriting}   & 31.20 & 1.49 & 0.79 \\
		DiffusionPen~\cite{nikolaidou2024diffusionpen} & 28.90 & 1.58 & 0.52 \\
		One-DM~\cite{one-dm2024}                & 15.60 & 0.75 & 0.53 \\
		HiGAN+~\cite{gan2022higan+}             & 10.17 & 0.44 & 0.33 \\
		\textbf{Ours-64}                       & \textbf{4.58} & \textbf{0.08} & \textbf{0.31} \\
		\bottomrule
	\end{tabular}
\end{table}

While global metrics like FID and KID measure overall similarity between real and generated handwriting distributions, they may not always capture writer-specific stylistic nuances. To gain a more detailed perspective, we perform a writer-level evaluation using FID, KID, and the Handwriting Distance (HWD) metric. For both the IAM and HANDS-VNOnDB datasets, we group real and generated samples by writer, compute each metric individually for every writer, and then average the results across all writers. This approach ensures that our evaluations focus on writer-specific consistency rather than being dominated by aggregate cross-writer statistics.

Table~\ref{tab:FID-KID-IAM-writer} presents the writer-level results on the IAM dataset. Our model (SpiS-GAN) achieves the lowest FID and KID scores at both resolutions: 4.37 / 0.06 (32px) and 4.58 / 0.08 (64px) — substantially better than the strongest competing baselines, FWGAN (FID: 6.73, KID: 0.22) and HiGAN+ (FID: 10.17, KID: 0.44). On the more challenging Vietnamese dataset HANDS-VNOnDB (Table~\ref{tab:FID-KID-VNOnDB-writer}), SpiS-GAN again surpasses all baselines with FID/KID scores of 3.67 / 0.14 (32px) and 4.37 / 0.19 (64px), reducing FID by nearly 63\% relative to HWT and by roughly 39\% relative to HiGAN+. These consistent improvements confirm that our model produces globally realistic handwriting distributions across different languages and resolutions.

However, as previously noted in subsection~\ref{sub:evaluation_procedure}, both FID and KID depend on an Inception network pre-trained on ImageNet, which creates a domain gap when applied to handwriting. To overcome this limitation, we employ the Handwriting Distance (HWD), a metric specifically designed to assess fine-grained stylistic consistency between real and generated handwriting. HWD uses a VGG16 network trained on a large handwriting corpus, enabling the feature embeddings to capture subtle handwriting characteristics.

When evaluated with this more handwriting-aware metric, SpiS-GAN continues to outperform all competing approaches. On the IAM dataset, it achieves HWD values of 0.51 (32px) and 0.31 (64px), compared to HiGAN+ (0.33) and One-DM (0.53) at 64 pixels. On HANDS-VNOnDB, SpiS-GAN reaches 0.37 (32px) and 0.20 (64px), outperforming HWT (0.62) and matching HiGAN+ (0.20). These findings confirm that even when assessed using a domain-specific, writer-sensitive metric, our model not only produces perceptually realistic handwriting (low FID/KID) but also excels at preserving stylistic coherence and writer identity (low HWD).

\begin{table}[!t]
	\centering
	\caption{Writer-level FID, KID, and HWD on the HANDS-VNOnDB dataset. Results marked with * are taken directly from FWGAN~\cite{TONGDANGKHOA2026130175}.}
	\label{tab:FID-KID-VNOnDB-writer}
	\begin{tabular}{lccc}
		\toprule
		\textbf{Method} & \textbf{FID} & \textbf{KID} & \textbf{HWD} \\
		\midrule
		\multicolumn{4}{l}{\textbf{32-pixel height}} \\
		\midrule
		VATr~\cite{pippi2023handwritten}       & 23.88 & 2.72 & 1.45 \\
		HiGAN~\cite{gan2021higan}              & 11.26 & 0.79 & 0.73 \\
		HWT~\cite{bhunia2021handwriting}       & 9.85  & 0.72 & 0.62 \\
		FWGAN~\cite{TONGDANGKHOA2026130175}        & 5.61  & 0.30 & 0.43 \\
		\textbf{Ours-32}   & \textbf{3.67} & \textbf{0.14} & \textbf{0.37} \\
		\midrule
		\multicolumn{4}{l}{\textbf{64-pixel height}} \\
		\midrule
		GANwriting*~\cite{kang2020ganwriting}   & 20.29 & 0.79 & 0.32 \\
		DiffusionPen*~\cite{nikolaidou2024diffusionpen} & 21.25 & 2.41 & 0.32 \\
		HiGAN+~\cite{gan2022higan+}             & 7.15  & 0.41 & 0.20 \\
		\textbf{Ours-64}                       & \textbf{4.37} & \textbf{0.19} & \textbf{0.20} \\
		\bottomrule
	\end{tabular}
\end{table}

\subsection{Model Size and Deployment Efficiency}

To evaluate the practicality of SpiS-GAN for real-world deployment, we measure its memory footprint and compare it against other handwriting synthesis models, as summarized in Table~\ref{tab:size}. Our framework requires a total of 54.50 MB (Generator: 47.91 MB, Encoder: 6.59 MB).

This places our model considerably below transformer-based architectures such as HWT (131.3 MB) and VATr (155.72 MB). It is marginally larger than FWGAN (52.96 MB), primarily due to the incorporation of our novel Star-Spiral Blocks in the generator. However, this slight increase is an intentional design choice. As discussed in Section~\ref{sec:generator}, the star operation offers an implicit expansion of feature space into higher dimensions without increasing the computational cost of the channel dimension. This enables our 47.91 MB generator to capture highly intricate handwriting styles that would typically demand a substantially larger network.

While HiGAN+ (21.7 MB) still holds the title of the most compact model, our framework delivers a substantial leap in visual quality (as evidenced by our FID/KID results) and downstream HTR performance. This demonstrates that our model achieves a highly favorable trade-off, remaining sufficiently lightweight for memory-constrained devices while attaining state-of-the-art handwriting generation capabilities.

\begin{table}[!t]
	\centering
	\caption{Model size comparison (Megabytes) focusing on Generator (Gen) and Encoder (Enc) modules.}
	\label{tab:size}
	\begin{tabular}{lccc}
		\toprule
		\textbf{Method} & \textbf{Gen (MB)} & \textbf{Enc (MB)} & \textbf{Total (MB)} \\
		\midrule
		HWT~\cite{bhunia2021handwriting}      & 80.7 & 50.6 & 131.3 \\
		VATr~\cite{pippi2023handwritten}      & 113.11 & 42.61 & 155.72 \\
		HiGAN~\cite{gan2021higan}             & 38.6 & 20.5 & 59.1 \\
		HiGAN+~\cite{gan2022higan+}           & \textbf{15.0} & 6.7 & \textbf{21.7} \\
		FWGAN~\cite{TONGDANGKHOA2026130175}             & 46.37 & \textbf{6.59} & 52.96 \\
		\textbf{Ours}                         & 47.91 & \textbf{6.59} & 54.50 \\
		\bottomrule
	\end{tabular}
\end{table}

	\subsection{Visual Comparison}

Metrics such as FID, KID, and HWD provide a solid quantitative assessment of overall quality, but they cannot fully capture the subtle nuances of human handwriting—such as connected strokes, natural slants, and consistent letter spacing. To better demonstrate how effectively our SpiS-GAN model preserves these fine details, we present a side-by-side visual comparison against state-of-the-art baselines.

We evaluate the models in two distinct settings: first, by generating entirely new text in a given style (generation quality); and second, by reconstructing the exact same text to assess how faithfully they replicate the original ink (reconstruction fidelity). The following examples highlight the key visual differences between our approach and existing methods.

\begin{figure}[!t]
	\centering
	\includegraphics[width=1\linewidth]{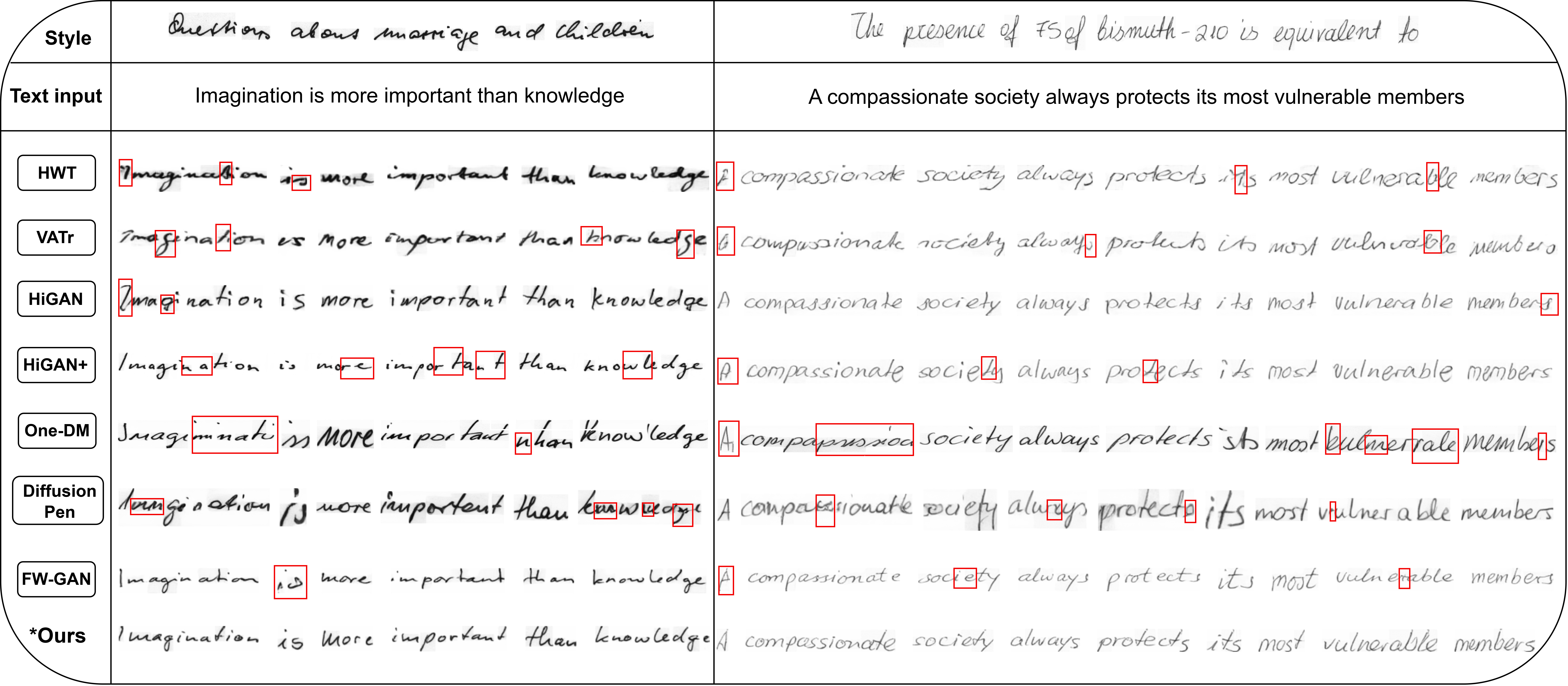}
	\caption{Qualitative comparison of handwriting generated from the IAM dataset across models, using identical style and content inputs. Each row corresponds to a different model, with red boxes highlighting failure cases (e.g., blurry or incorrect characters).}
	\label{fig:qualitative_gen_eng}
\end{figure}

\begin{figure}[!t]
	\centering
	\includegraphics[width=0.9\linewidth]{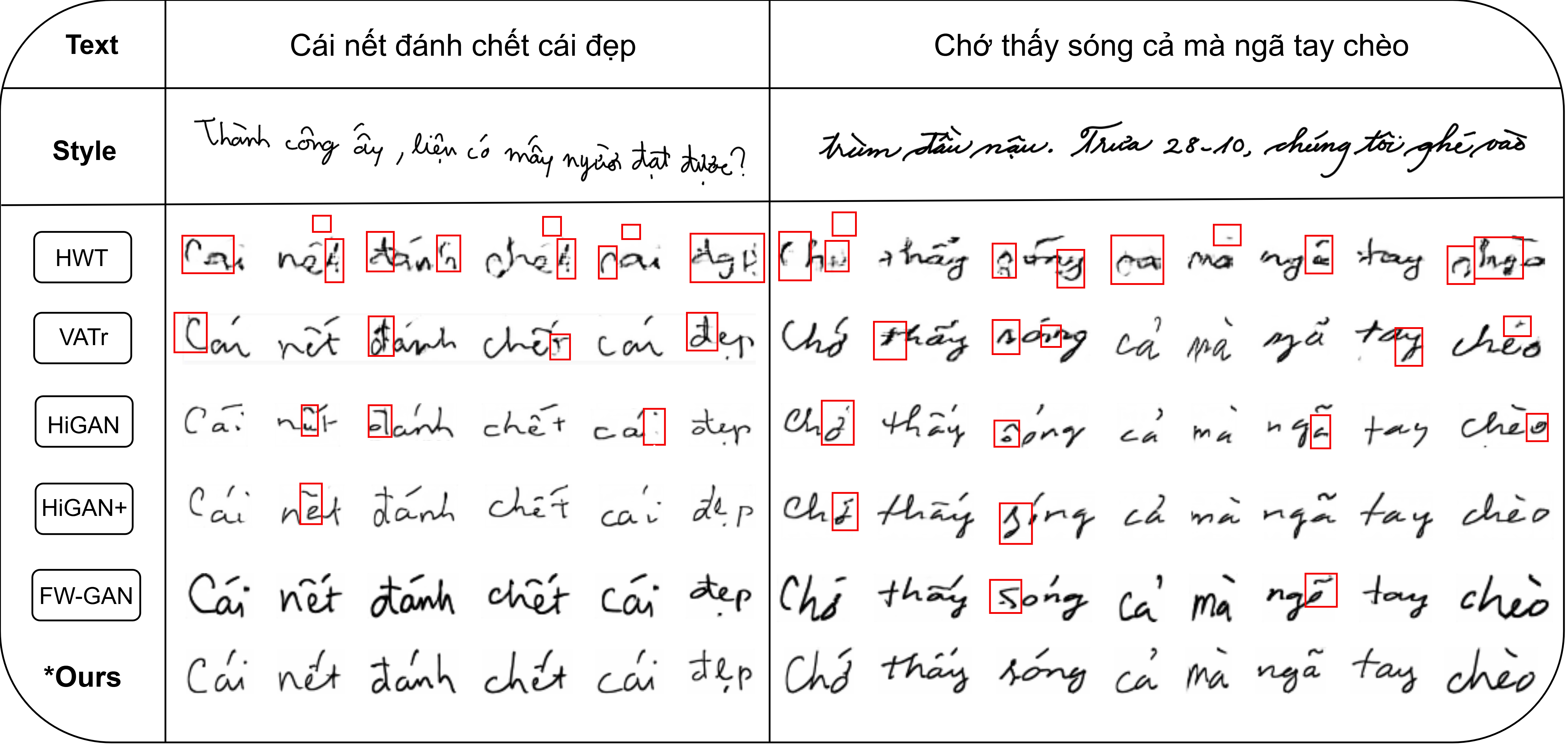}
	\caption{Qualitative comparison of handwriting generated from the HANDS-VNOnDB dataset across models, using identical style and content inputs. Each row corresponds to a different model, with red boxes highlighting failure cases (e.g., blurry or incorrect characters).}
	\label{fig:qualitative_gen_vietnamese}
\end{figure}

\subsubsection{Generation Quality}

Figures~\ref{fig:qualitative_gen_eng} and \ref{fig:qualitative_gen_vietnamese} offer a visual comparison of handwriting produced by our model against recent baselines, including HWT~\cite{bhunia2021handwriting}, VATr~\cite{pippi2023handwritten}, HiGAN~\cite{gan2021higan}, HiGAN+~\cite{gan2022higan+}, DiffusionPen~\cite{nikolaidou2024diffusionpen}, One-DM~\cite{one-dm2024}, and FW-GAN~\cite{TONGDANGKHOA2026130175}. We exclude GANWriting~\cite{kang2020ganwriting} due to the lack of a compatible checkpoint. One-DM and DiffusionPen are evaluated only on IAM (English), while the remaining models are tested on both IAM and HANDS-VNOnDB (Vietnamese). Red boxes indicate artifacts produced by the baselines. In contrast, our framework (\textit{*Ours}) consistently generates legible, stylistically accurate, and artifact-free text across both languages.

On the IAM dataset, existing approaches show clear weaknesses in shape stability and character fidelity. HWT frequently drops strokes (e.g., the disconnected ``I''), while VATr severely distorts complex letter forms (e.g., deformed ``g'' and ``k''). HiGAN produces malformed uppercase letters and introduces noise at word endings. HiGAN+ creates unnatural spacing between letters, breaking the cursive flow and producing disjointed characters. One-DM exhibits severe spelling errors, completely scrambling the characters in words like ``Imagination'' and ``compassionate''. DiffusionPen preserves the overall style but suffers from noticeable ink bleeding, turning sharp loops into blurred smudges. Even FW-GAN, the previous state-of-the-art, displays localized flaws, such as distorted letters (``is'', ``A'', ``e'') and a missing stroke in the ``r'' of ``vulnerable''. Our model addresses these limitations effectively. By applying strict structural constraints, it maintains accurate stroke thickness and consistent natural slant, resulting in sharper handwriting.

The HANDS-VNOnDB dataset introduces extra complexity through its composite characters and stacked diacritical marks. HWT and VATr fail to render essential features like the crossed ``d'' and produce fragmented outputs. HiGAN and HiGAN+ maintain better stroke continuity but frequently misplace or omit diacritics, including circumflexes and tildes. FW-GAN achieves notable improvement but still distorts certain loops (e.g., ``s'') and generates malformed tildes. Moreover, FW-GAN struggles to replicate the exact stroke thickness of the reference, often producing overly heavy text. With the MLP-augmented Discriminator and our proposed loss functions, our model seamlessly overcomes these issues. It accurately places all diacritics and faithfully reproduces the reference ink density and continuous flow.

\begin{figure}[!t]
	\centering
	\includegraphics[width=0.75\linewidth]{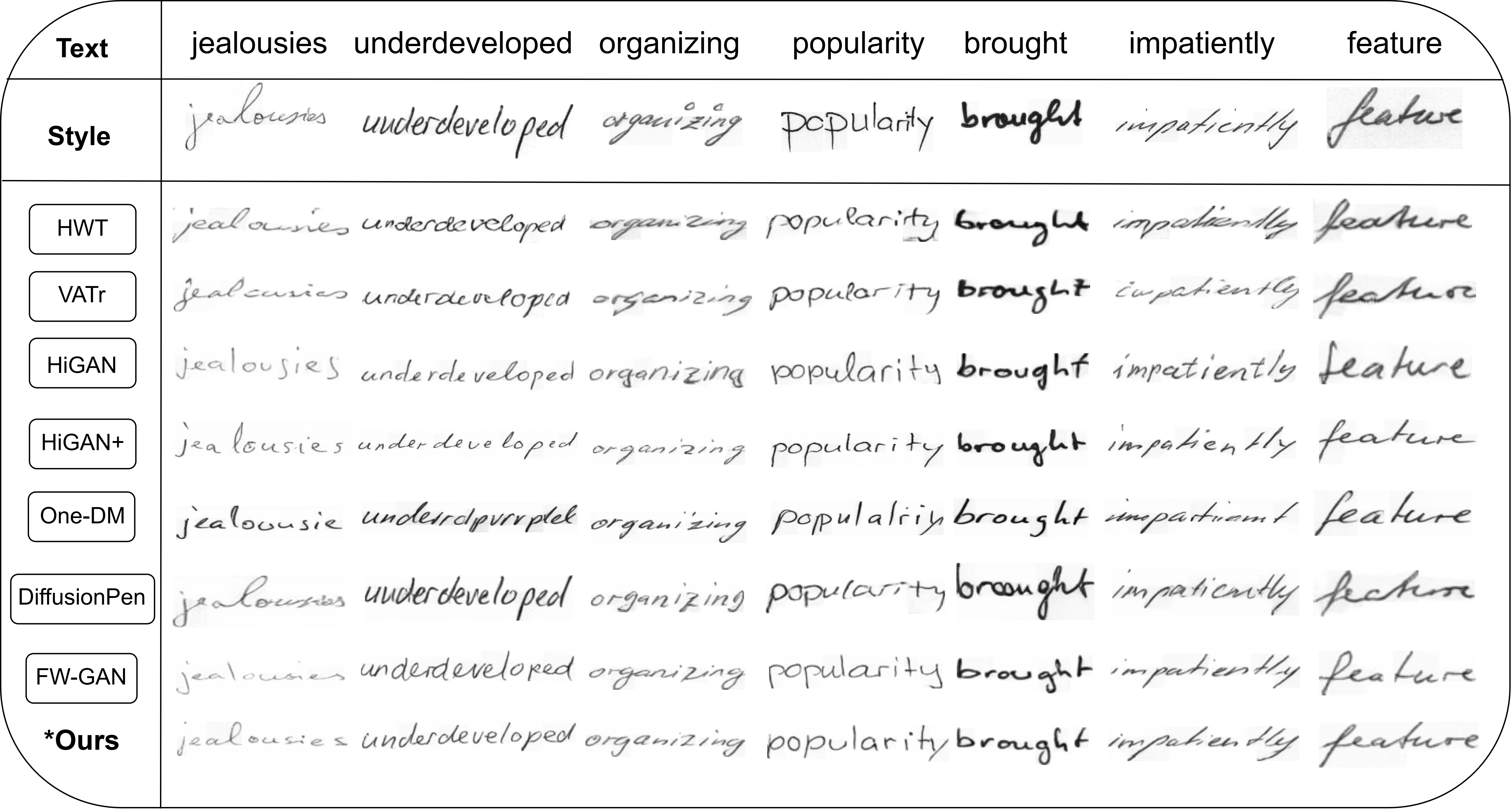}
	\caption{Qualitative reconstruction results. Each row corresponds to a different model, conditioned on the same ground-truth style (top row) and target text (bottom row). The goal is to reproduce the target content while preserving the handwriting style.}
	\label{fig:reconstruction}
\end{figure}

\subsubsection{Reconstruction Quality}

Figure~\ref{fig:reconstruction} illustrates the reconstruction performance, where models are required to reproduce the target text in the reference style. Earlier baselines such as HWT and VATr produce faint, fragmented strokes. HiGAN and HiGAN+ struggle with inter-letter spacing, often breaking continuous words into disconnected pieces (visible in ``jea lousies'' and ``under deve loped''). One-DM fails to preserve semantic content, generating random characters instead of the target spelling. DiffusionPen shows severe ink smudging, blurring thick strokes in words like ``brought'' and ``feature''. Although FW-GAN performs competitively, it struggles with extreme stylistic variations, failing to match the heavy marker ink in ``brought'' or the ultra-thin strokes in ``impatiently''. In contrast, our model captures this full range with remarkable precision. Through the integration of two novel blocks in the generator and our proposed loss functions, it faithfully reproduces everything from faint pencil lines to bold ink strokes, without fragmentation, hallucinations, or smudging, demonstrating superior fine-grained calligraphic control.

\section{Conclusion}\label{sec:conclusion}
In this paper, we introduce SpiS-GAN, a novel GAN-based model designed for one-shot handwriting synthesis, which tackles the problem of producing realistic handwritten text that maintains stylistic consistency from only one reference example. By combining Modulated Elliptical SpiralFC (MESpiralFC) with the star operation within our Star-Spiral Blocks (SSB), the generator overcomes the limitations of conventional grid-based architectures and enables implicit high-dimensional feature expansion to enable rich feature aggregation. A Spiral-Modulated discriminator, paired with a Sobel-Regularized Edge Reconstruction Loss (SELoss), effectively preserves minimal details and prevents the loss of fine edge details
through explicit edge supervision. Extensive experiments on English and Vietnamese benchmarks confirm that SpiS-GAN consistently surpasses state-of-the-art methods, producing highly authentic, style-preserving outputs while significantly boosting downstream handwriting recognition performance. To the best of our knowledge, this work is probably the first to utilize the application of the StarNet-style multiplication to offline handwriting synthesis for efficient complex feature interactions, together with the introduction of MLP-augmented modules in the discriminator. Our framework advances the field of handwriting synthesis while offering a robust solution for HTR data augmentation and personalized text generation.

\section*{Acknowledgements}
This research was supported by The VNUHCM-University of Information Technology’s Scientific Research Support Fund.

\bibliographystyle{elsarticle-num}
\bibliography{references} 

\end{document}